\documentclass[11pt]{revtex4}
\usepackage{amsmath,amsthm,amssymb}
\usepackage{amsfonts,amssymb}
\usepackage{graphicx}
\usepackage{mathtools}
\usepackage[colorlinks=true, pdfstartview=FitV, linkcolor=blue,citecolor=blue, urlcolor=blue]{hyperref}
\usepackage{bm}
\usepackage{algorithm}
\usepackage{algpseudocode}

\numberwithin{equation}{section}

\makeatletter
\renewcommand*\env@matrix[1][*\c@MaxMatrixCols c]{%
  \hskip -\arraycolsep
  \let\@ifnextchar\new@ifnextchar
  \array{#1}}
\renewcommand\Dated@name{}  
\renewcommand*{\thesection}{\arabic{section}}

\renewcommand*{\p@subsection}{}

\renewcommand*{\p@subsubsection}{}
\makeatother

\newcommand{\be}{ \begin{equation} }
\newcommand{\ee}{ \end{equation} }

\usepackage{xcolor}

\begin{document}

\title{Multi-stage Neural Networks:  Function Approximator of Machine Precision}
\date{} %
\author{Yongji Wang \footnote{Corresponding author: yongjiw@stanford.edu}}\vspace{0.2em}
\affiliation{Department of Geosciences, Princeton University, Princeton, NJ 08544}
\affiliation{Department of Geophysics, Stanford University, Stanford, CA 94305}

\author{Ching-Yao Lai \footnote{Corresponding author: cylai@stanford.edu}} \vspace{0.2em}
\affiliation{Department of Geophysics, Stanford University, Stanford, CA 94305}
\affiliation{Department of Geosciences, Princeton University, Princeton, NJ 08544}

\begin{abstract}
Deep learning techniques are increasingly applied to scientific problems, where the precision of networks is crucial. Despite being deemed as universal function approximators, neural networks, in practice, struggle to reduce the prediction errors below $O(10^{-5})$ even with large network size and extended training iterations. To address this issue, we developed the multi-stage neural networks that divides the training process into different stages, with each stage using a new network that is optimized to fit the residue from the previous stage. Across successive stages, the residue magnitudes decreases substantially and follows an inverse power-law relationship with the residue frequencies. The multi-stage neural networks effectively mitigate the spectral biases associated with regular neural networks, enabling them to capture the high frequency feature of target functions. We demonstrate that the prediction error from the multi-stage training for both regression problems and physics-informed neural networks can nearly reach the machine-precision $O(10^{-16})$ of double-floating point within a finite number of iterations. Such levels of accuracy are rarely attainable using single neural networks alone. \\
\\
{\bf keywords}: scientific machine learning; neural networks; physics-informed neural networks, multi-stage training
\end{abstract}

\maketitle

\renewcommand\theequation{\arabic{section}.\arabic{equation}}

\section{Introduction}
Deep learning techniques \cite{lecun2015deep} have been well developed in the fields of computer vision \cite{RFB15a, mildenhall2021nerf} and natural language processing \cite{collobert2008unified,devlin2018bert,chowdhary2020natural}. More recently, neural networks have been increasingly applied to the mathematical and physical sciences \cite{kochkov2021machine,lemos2022rediscovering,wang2022self}, where the demand for precision is high. In particular, physics-informed neural networks (PINNs) \cite{raissi2019physics, karniadakis2021physics} have emerged as a new class of numerical solver for partial differential equations, where computing high-precision solutions becomes an intrinsic requirement of the method.

Neural networks have been proven to be universal function approximators \cite{hornik1989multilayer, hornik1990universal}. However, in practice, neural network training often falls into local minima \cite{baldi1989neural,krishnapriyan2021characterizing}, causing the training loss to plateau after a certain number of iterations $n_{iters}$. This issue cause the failure modes of PINNs \cite{krishnapriyan2021characterizing}. Advanced methods focusing on different aspects, such as activation function selection \cite{sitzmann2020implicit, saragadam2023wire}, network configuration \cite{he2016deep, jagtap2021extended, moseley2021finite}, optimization techniques \cite{tu2019autozoom,chiu2022can}, trainable weights \cite{mcclenny2020self}, and loss function \cite{li2021autobalance, van2022optimally}, have been developed to effectively enhance the convergence rate of the loss function for various problems. However, few of these methods manage to reduce the training error less than $O(10^{-5})$. In contrast, classical numerical methods (e.g., finite difference) can systematically enhance solution's accuracy by simply reducing the grid size \cite{ralston2001first}. This is a major shortcoming of neural networks for solving many problems within mathematical and physical sciences.

In this work, we proposed the multi-stage neural networks that effectively addresses this limitation. Our novel method involves dividing the network training into multiple stages, where each stage incorporates a separate neural network. The setting of each network in a given stage are optimized based on the residues from the preceding stage. By executing training stage by stage, we significantly enhance the convergence rate, ensuring that it remains consistently high throughout the iterations. As a result, the combined neural networks from different stages can approximate the target function with remarkable accuracy, with the error approaching the machine precision $O(10^{-16})$ for double-floating point numbers.

We begin with the introduction of the multi-stage neural network for regression problems in Section \ref{sec:reg}. By exploring the limitations of classical neural network training, we highlight the benefits of the multi-stage training in overcoming these constraints. We then propose and substantiate the optimal settings for each training stage. In Section \ref{sec:PINN}, we extend the method to physics-informed neural networks (PINNs) for solving differential equations. Unlike regression problems, the optimal settings for PINNs in each stage are implicitly tied to the equation residues of previous stages. Both theoretical investigation and practical algorithmic solutions are presented to address this challenge. Additional techniques that can expedite the multi-stage training for PINNs are also discussed. In Section \ref{sec:combPINN}, we generalize the multi-stage training scheme to solve combined-forward-and-inverse problems, which are of great importance in mathematical and physical sciences. Lastly, we provide further discussions on the challenges and potential development of the MSNN method in Section \ref{sec:dissc} and conclude the paper in Section \ref{sec:concl}.

\section{Multi-stage training scheme for regression problems}\label{sec:reg}
We first illustrate the multistage idea with regression problems that involve predicting a continuous output variable $u$ as a function of the input variable $x$. We train a neural network that represents $u(x)$ to fit $N_d$ data points, denoted $(x_i, u_i)$. The loss function for a regression problem is typically the mean squared error (MSE), defined as
\begin{gather}\label{eq:lossPINN}
	\mathcal{L} = \frac{1}{N_d} \sum_{i=1}^{N_d} [{u}(x_i) - u_i]^2 \, ,
\end{gather}
In this study, we consider all training data with no noise.

\subsection{Limitation of regular neural network training} \label{sec: fit}
\begin{figure}
	\centering
	\includegraphics[width=1\textwidth]{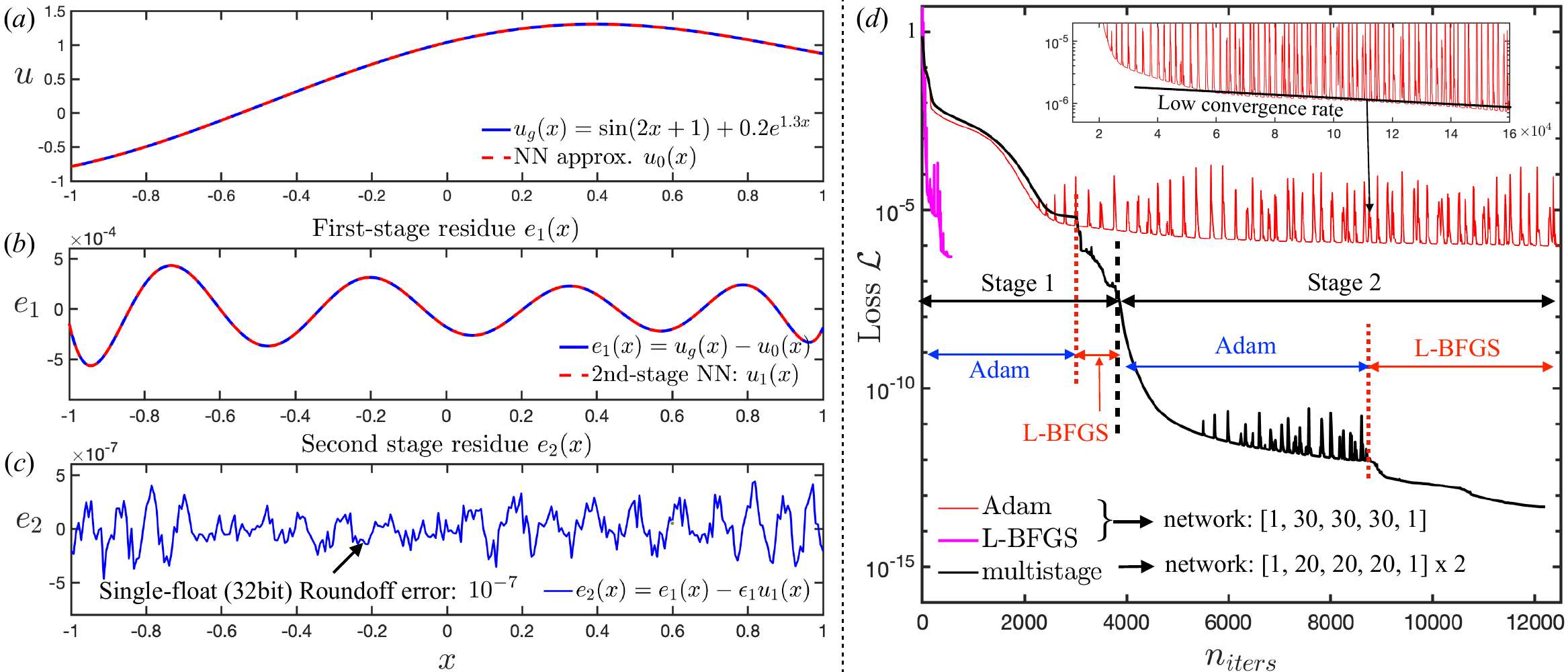}
	\caption{\label{fig:RegmNN} {\bf Comparison of single-stage with multi-stage training}. ($a$) Fitting of a neural network $u_0(x)$ with \texttt{tanh} activation function to the data from \eqref{eq:eqn1}. ($b$) Fitting of second-stage neural network to the error $e_1(x)$ between the data from \eqref{eq:eqn1} and the first-stage trained network $u_0(x)$ as shown in ($a$). ($c$) The error $e_2(x)$ between the data and the sum of two-stage networks, which reaches the machine precision of a single float (32-bit). ($d$) Comparison of the loss convergence between a single-stage training (pink and red) and a two-stage training (black). For a single-stage training, the convergence rate of loss suddenly reduces (for Adam) after the loss reaches $O(10^{-6})$ or terminates (for L-BFGS). For a two-stage training, even with less number of weights and biases, the convergence rate is significantly faster than that for the single-stage training.}
\end{figure}

To illustrate the limitation of neural network's function approximation capacity, we consider a target function,
\begin{equation} \label{eq:eqn1}
	u_g(x) = \sin(2x+1) + 0.2 e^{1.3x} \, ,
\end{equation}
we created training data by sampling 300 data points from it with no noise, uniformly distributed within the domain $x \in [-1, 1]$. To fit the training data, we create a fully-connected neural network made of three hidden layers with 20 units in each layer and use hyperbolic tangent as the activation function for each unit. Using Adam \cite{kingma2014adam} optimizer, figure \ref{fig:RegmNN}($a$) shows that the trained neural network $u_0(x)$ captures the target function $u_g(x)$ well. During the iterations, the training loss $\mathcal{L}$ based on mean squared error (MSE) between the data and network, decreases significantly at the early stage (figure \ref{fig:RegmNN}$b$). However, after 5000 iterations, it reaches a plateau around $O(10^{-7})$ with very small convergence rate. The error function $e_1(x)$ between $u_g$ and $u_0$ across the training domain is, thus, trapped around $10^{-4}$ (inset of figure \ref{fig:RegmNN}$a$). Further experiments, elucidated in \ref{sec:AppA}, affirm that this plateau value of the error remains consistent even with larger networks and additional data, and not optimizer-specific.

Neural networks are known for their spectral biases \cite{rahaman2019spectral}, also referred to as the frequency principle \cite{xu2019frequency}. Utilizing the tool of neural tangent kernels \cite{jacot2018neural}, prior studies \cite{tancik2020fourier, wang2021eigenvector} demonstrated that a standard multi-layer neural network struggles to learn the high frequencies of target functions in both theory and practice. The plateau of training loss corresponds to a mismatch between the trained network $u_0(x)$ and target function $u_g(x)$ at high frequencies. Figure \ref{fig:RegmNN}($b$) demonstrates that the error function $e_1(x) = u_g(x) - u_0(x)$ within the domain is indeed a high-frequency function.

\subsection{Key settings of multi-stage training scheme}\label{sec:mNN_reg}
Since training a single neural network struggles with learning the high frequencies of the target function, an intuitive approach is to train a second neural network to capture the error function $e_1(x)$, or the residue, between the training data and the first trained network \cite{michaud2023precision}. The original training data from \eqref{eq:eqn1} is denoted with $(x^{(i)}, u_g^{(i)})$. The training data for the second neural network would be $(x^{(i)}, e_1(x^{(i)}))$, where $e_1(x^{(i)})$ denotes the error of network at $x^{(i)}$. At this point, extra care should be taken when setting up the second neural network, particularly concerning two key aspects.

\subsubsection{Magnitude of the second neural network}
Considering that the original training data has a magnitude of $O(1)$, then the training data for the second neural network, which is the residue $e_1(x)$, would be much smaller than 1 (figure \ref{fig:RegmNN}$b$). We observe that a neural network employing regular weight initialization methods, such as Xavier \cite{glorot2010understanding}, often struggles to capture training data whose magnitude is significantly larger or smaller than 1 (see \ref{sec:AppB}). A straightforward solution to this issue is to normalize the training data by its root mean square value $\epsilon_1$, defined as
\begin{eqnarray}
	\epsilon_1 = \sqrt{\frac{1}{N_d}\sum_{i=1}^{N_d} [e_1(x^{(i)})]^2} = \sqrt{\frac{1}{N_d}\sum_{i=1}^{N_d} [u_g^{(i)}-u_0(x^{(i)})]^2 } \, .
\end{eqnarray}
 Then, the normalized training data for the second neural network becomes $(x^{(i)},  e_1(x^{(i)})/\epsilon_1)$. Denoting the second trained network as $u_1(x)$, the combined networks for the original data become
 \begin{equation}
 	u_c^{(1)}(x) = u_0(x) + \epsilon_1 u_1(x)\, .
 \end{equation}
Subsequently, we can continue training the third or even further neural networks to reach higher accuracy for our model. The training data for the $(n+1)$-th neural network $u_n$ is the residue $e_n$ between the original training data $u_g$ and the output of the previously combined $n$ neural networks, $u_c^{(n-1)}(x^{(i)})$, normalized by its own magnitude (root mean square value) $\epsilon_n$, namely $(x^{(i)},  e_n(x^{(i)})/\epsilon_n)$. Then, the final model that combines all the $(n+1)$ neural networks reads,
 \begin{equation}
	u_c^{(n)}(x) =  \sum_{j=0}^{n} \epsilon_{j} u_{j}(x)\, ,
\end{equation}
where $\epsilon_i$ stands for the magnitude for the $i$-th neural network. When the original training data $u_g$ is normalized, $\epsilon_0$ is set to be 1.

\begin{figure}
	\centering
	\includegraphics[width=1\textwidth]{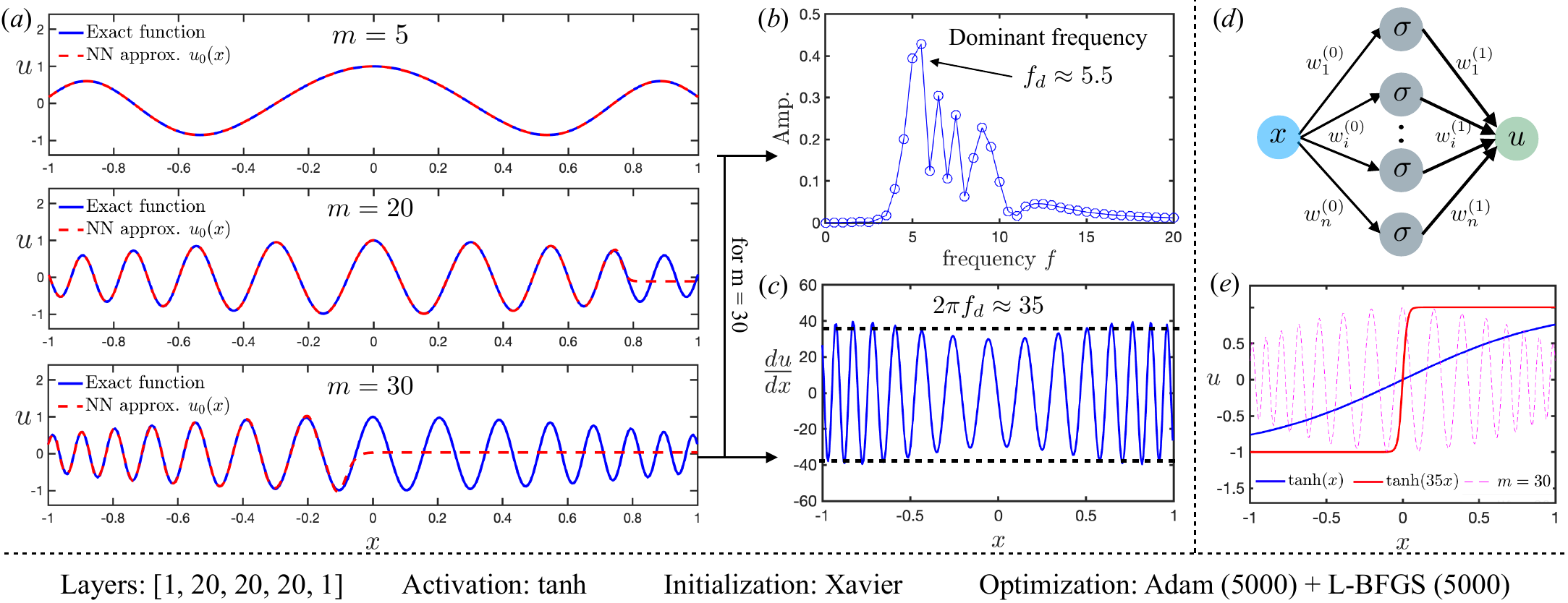}
	\caption{\label{fig:freq1} {\bf Spectral biases of neural networks}. ($a$)  Fitting of neural networks with \texttt{tanh} activation function to the data from \eqref{eq:eqn2} for different $m$. Under regular settings, neural networks have difficulty fitting high-frequency functions. ($b$) Frequency domain of the function \eqref{eq:eqn2} for $m=30$ with the dominant frequency $f_d = 5.5$. ($c$) Derivative of the function \eqref{eq:eqn2} $du/dx$ for $m=30$, which scale as $O(2\pi f_d)$. ($d$) Schematic diagram of a single-hidden layer neural network. ($e$) Comparison between single-neuron outputs for different weights $w^{(0)}$ within the \texttt{tanh} activation function and the function \eqref{eq:eqn2} for $m=30$. To capture high-frequency functions, it shows that the weight within the activation function needs to increase from $O(1)$ to $O(2\pi f_d)$. }
\end{figure}

\subsubsection{Frequency of the second neural network} \label{sec:freq_reg}
Even with normalization, the second neural network, if initialized with regular weights, could still struggle to fit the high-frequency data due to the inherent spectral biases of neural networks.  To illustrate this, we consider a target function 
\begin{equation} \label{eq:eqn2}
	u(x) = \left(1-\frac{x^2}{2}\right)\cos\left[m\left(x + 0.5x^3\right)\right].
\end{equation}
with $m$ the free parameter related to the frequency of the function.  Figure \ref{fig:freq1} shows the function \eqref{eq:eqn2} for $m = 3$, 15, and 30, respectively. For each $m$, we generate 300 sample points $(x_i, u_i)$ that satisfy \eqref{eq:eqn2} as our training data, with $x_i$ uniformly distributed in the domain $[-1, 1]$. Figure \ref{fig:freq1} shows that the neural network, using regular weight initialization, fits the data well for $m=3$, partially misses the data for $m=15$, and completely fails to fit the data for $m=30$.

\begin{figure}
	\centering
	\includegraphics[width=1\textwidth]{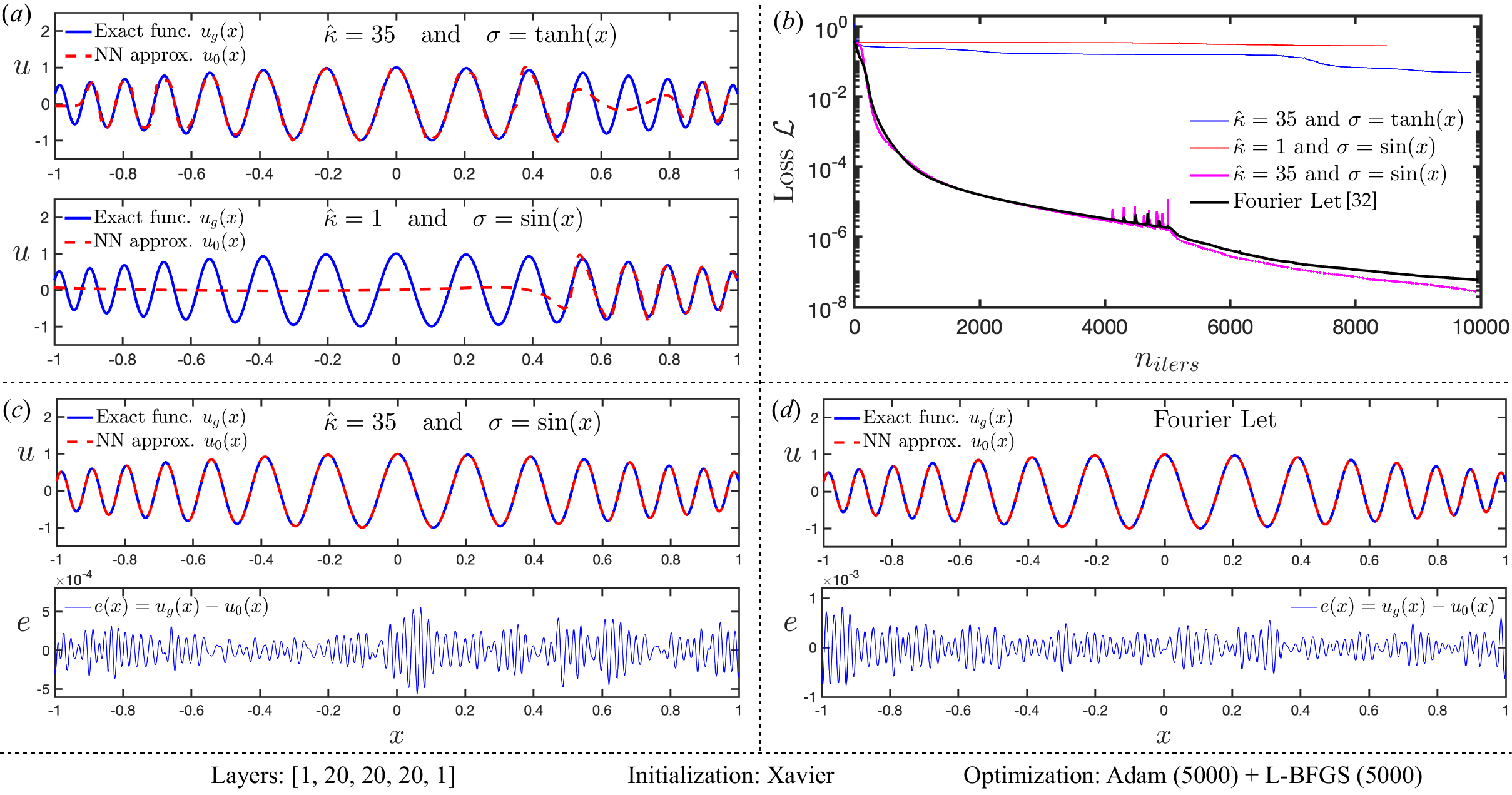}
	\caption{\label{fig:freqSet} {\bf Neural network settings for high-frequency functions}. ($a$)  Fitting of a neural network to the data from \eqref{eq:eqn2} with $m = 30$, by either changing the activation function for the first hidden layer to $\sin(x)$, or multiplying the weight $w_i^{(0)}$ before the first hidden layer by a scale factor $\kappa$. None of them captures the data. ($b$) Comparison of training loss for the neural networks with different settings. ($c$) Neural network using $\sin(x)$ activation function and modified scale factor $\hat{\kappa} = 35$ fits well the high-frequency data \eqref{eq:eqn2} with $m = 30$, reaching the same accuracy as ($d$) Fourier Let network \cite{tancik2020fourier}. }
\end{figure}

To understand the challenge in fitting high-frequency data, let's consider a shallow neural network with a single input, single output, and one hidden layer that uses the hyperbolic tangent as its activation function:
\begin{equation}\label{eq:singleNN}
	u(x) = \sum_{i=1}^N w_{i}^{(1)} \: \tanh\left(w_{i}^{(0)}x + b_{i}^{(0)} \right) +b_0 \, ,
\end{equation}
where $w_i^{(0)}$ denote the weights between the input and hidden layers, and $w_i^{(1)}$ are the ones between the hidden and output layers. $b_{i}^{(0)}$ is the bias for the hidden units and $b_0$ is the bias for the output unit.  The magnitude of the output function is determined by $w_i^{(1)}$, while $w_i^{(0)}$, within the activation function, influence the local gradient of the function (figure \ref{fig:freq1}$e$).  Common practice involves initializing the weights of the network to follow a Gaussian distribution with zero mean and a specified variance. For example, Xavier initialization uses a specified variance $V_{ar} = \sqrt{2/(N_{l-1} + N_{l})}$, where $N_{l-1}$ and $N_{l}$ are the number of units in the preceding and succeeding layers, respectively. This initialization ensures that the variance of the sum of all unit outputs in each hidden layer remains $O(1)$, which prevents gradient vanishing or explosion during the training. However, a side effect of this approach is that the neural network becomes a slowly varying function with respect to normalized inputs.

For a high-frequency function with normalized input and output, and a dominant frequency $f_d$, the magnitude of its gradient scales as $O(2\pi f_d)$ (figure \ref{fig:freq1}$c$). To capture these large gradients, considering a one-hidden layer network with single input and output \eqref{eq:singleNN}, the weights $w_i^{(0)}$ within the activation function need to increase from their initialized value of $O(\sqrt{1/V_{ar}})$ to $O(2\pi f_d)$ during training. This large shift in weight values, particularly for large $f_d$, leads to slower convergence during training or an inaccurate approximation of the data.

\begin{figure}
    \centering
    \includegraphics[width=1\textwidth]{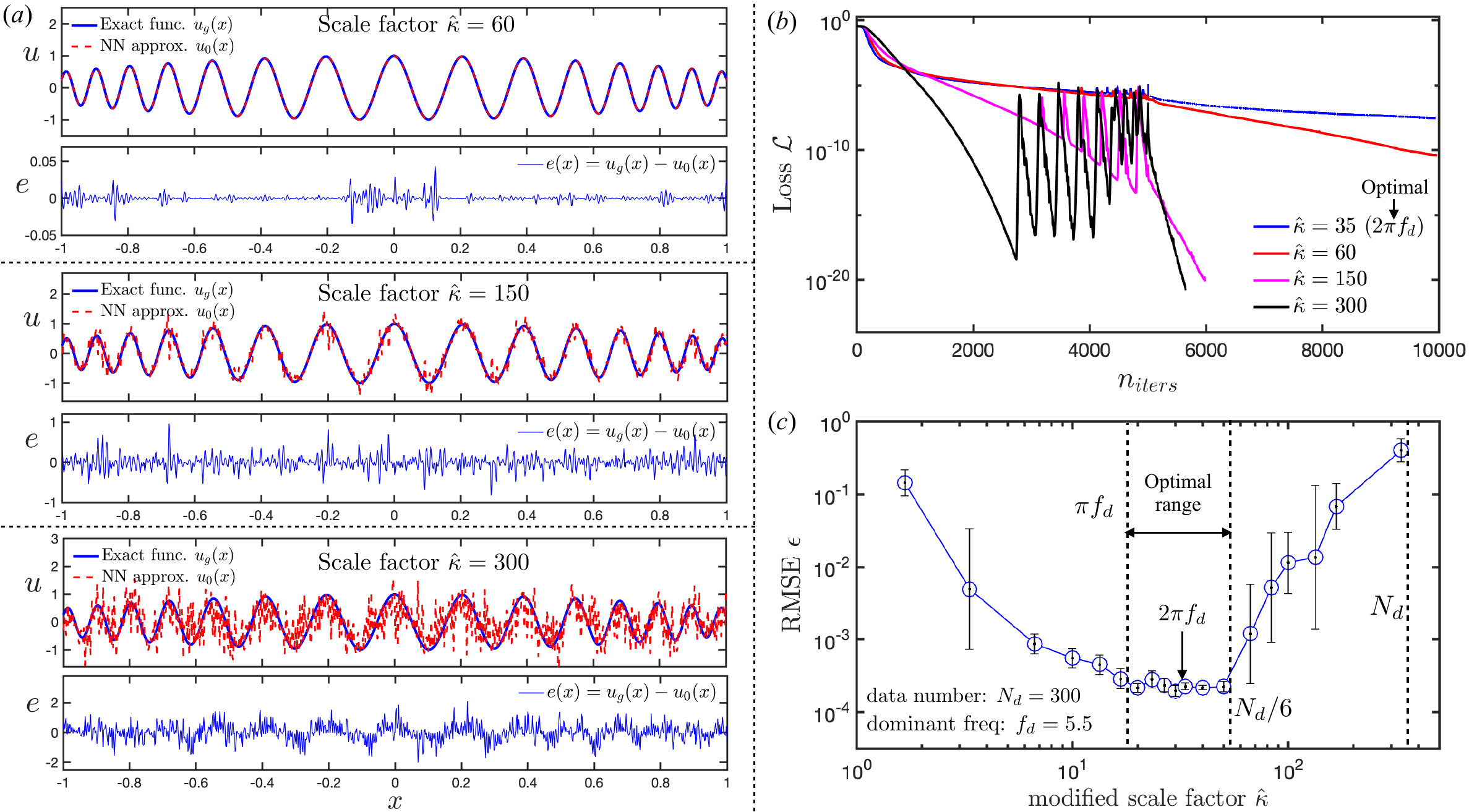}
    \caption{\label{fig:overfit} {\bf Importance of the modified scale factor $\hat{\kappa}$.} ($a$)  Fitting of neural networks to the data from \eqref{eq:eqn2} for $m = 30$ using different modified scale factor $\hat{\kappa}$. The networks start overfitting the data when $\hat{\kappa} \geq 60$. ($b$) Training loss for the neural networks with different modified scale factors $\hat{\kappa}$. When $\hat{\kappa} \geq 60$, the training loss decreases significantly fast due to over-fitting. ($c$) Relation of the root mean square value $\epsilon$ of the error $e(x)$ between the trained network $u(x)$ and target function $u_g(x)$ with the modified scale factor $\hat{\kappa}$. The minimal error is reached when $\pi f_d < \hat{\kappa} < N_d/6$, where $f_d$ denotes the dominant frequency and $N_d$ the total number of training data points. }
\end{figure}

To address this issue, we multiply weights within the activation function by a large scale factor $\kappa$ \cite{jagtap2020adaptive} to expedite the convergence of weights towards their optimal high value when fitting high-frequency data. We only multiply the scale factor $\kappa$ to the weights between the input and the first hidden layer, rather than all weights, to prevent gradient explosion \cite{aggarwal2018neural} during the training. 

Besides large gradient, high-frequency functions also have a large amount of inflection points. In contrast, the hyperbolic tangent function, being a monotonic function, struggles to capture this feature. Periodic functions, such as the sine or cosine function, are more suitable choices for activation functions in this case \cite{sitzmann2020implicit}.  In our approach, we use the sine function solely for the first hidden layer while retaining the hyperbolic tangent function for the remaining layers. This combination allows us to capture both low and high-frequency data effectively. 

Figure \ref{fig:freqSet} illustrates the impact of the scale factor $\kappa$ and the choice of activation function on improving the fit for high-frequency data. Using a combination of the scale factor $\kappa$ and sine function for the first hidden layer yields the best training result. This combination equates to applying a Fourier feature mapping (i.e. Fourier let network) \cite{tancik2020fourier} to the input before it is passed through the multi-layer network.  Figure \ref{fig:freqSet}($c\, \& \, d$) compares the convergence rate and final error of both methods when fitting the same high-frequency data. The results are consistently good, verifying the efficacy of both methods in fitting high-frequency data.

To expedite the convergence of weights from their initialized value $O(\sqrt{1/V_{ar}})$ to the high gradient value $O(2\pi f_d)$ of a high frequency function, the optimal value of $\kappa$ is expected to depend on the variance $V_{ar}$, which is relevant to the weight initialization approaches, and the size of neural network. To isolate the impact of $V_{ar}$ on determining the optimal value of the scale factor, we introduce a modified scale factor $\hat{\kappa}$,
\begin{equation}
	\hat{\kappa} = \kappa /\sqrt{V_{ar}} \, ,
\end{equation}
Figure \ref{fig:overfit}($c$) shows that the minimal fitting error is achieved when the modified scale factor is
\begin{equation} \label{eq:sfc}
    \hat{\kappa} > \pi  f_d \, ,  \qquad \text{namely} \qquad \kappa > \pi f_d \sqrt{V_{ar}}\, ,
\end{equation}
where $f_d$ denotes the dominant frequency of the data. This finding is intuitive, as a scale factor $\hat{\kappa}$ that meets the criterion \eqref{eq:sfc} allows the neural network to directly capture the large gradient $O(2\pi f_d)$ of the high-frequency data. 

However, setting $\hat{\kappa}$ too high, close to the number of data points $N_d$, results in overfitting of the neural network.  Figure \ref{fig:overfit}($a$) shows that a neural network trained with a scale factor $\hat{\kappa} = 300$ to fit the training data ($N_d=300$) sampled from the high-frequency function \eqref{eq:eqn2} with $m=30$ overfits the data. While the training loss is significantly small (figure \ref{fig:overfit}$b$), the validation error is extremely large. To mitigate overfitting, figure \ref{fig:overfit}($c$) suggests that the modified scale factor $\hat{\kappa}$ should be less than one-sixth of the total number of data points $N_d$. As a rule of thumb, for the optimal fitting of high-frequency data, besides satisfying \eqref{eq:sfc}, the number of training data points $N_d$ should also meet the criterion
\begin{equation}\label{eq:nscri}
	N_d/6  > \pi f_d \qquad \Longrightarrow \qquad N_d > 6\pi f_d.
\end{equation}
Given that the training domain is often normalized within $[-1, 1]$, which contains $2f_d$ dominant periods, the criterion \eqref{eq:nscri} essentially requires a minimum of $3\pi \approx 10$ data points within each dominant period $1/(2f_d)$, to ensure optimal fitting of the neural network to the high-frequency data. Without specific clarification, the criterion \eqref{eq:nscri} is applied to all the example problems in this section. 

\subsection{Algorithm of multi-stage training scheme for regression problems}\label{sec: mNN}
Incorporating these key settings for higher-stage neural network training, we summarize a complete procedure of multi-stage training scheme for regression problems as shown in Algorithm \ref{alg:cap}.

\begin{algorithm}[h]
	\caption{Multi-stage training scheme for regression problems}\label{alg:cap}
	\begin{algorithmic}[1]
	\State Normalizing both the input and output data.  \vspace{0.2em}
	\item Building the first-stage neural network with regular weight initialization. \vspace{0.2em}
	\item Training the neural network to fit the normalized data. \vspace{0.2em}
	\item Obtaining the output of the trained neural network $u_0(x)$. \vspace{0.2em}
	\item Calculating the error $e_1(x) = u_g - u_0$ between the data $u_g$ and the trained network $u_0(x)$. \vspace{0.2em}
	\item {\bf} Normalize the error $e_1(x)$ by its root mean square value $\epsilon_1$ 
    \item Building the second neural network with the scale factor $\kappa$ obtained from the dominant frequency $f_d$ of the error $e_1(x)$. \vspace{0.2em}
    \item Training the neural network to fit the normalized error $e_1(x)/\epsilon_1$.  \vspace{0.2em}
	\item Repeating Step 4-9 for certain times until $e_{n+1} = e_{n} - \epsilon_n u_{n}$ is smaller enough. \vspace{0.2em}
	\item Generating the final model, $u(x) = \sum_{i=0}^{n} \epsilon_i u_{i}(x)$, by combining all trained neural networks at different stages $u_{n}$. This final model can approximate the original data $u_g$ with exceptionally high accuracy. 
\end{algorithmic}
\end{algorithm}

 \begin{figure}[t]
	\centering
	\includegraphics[width=0.99\textwidth]{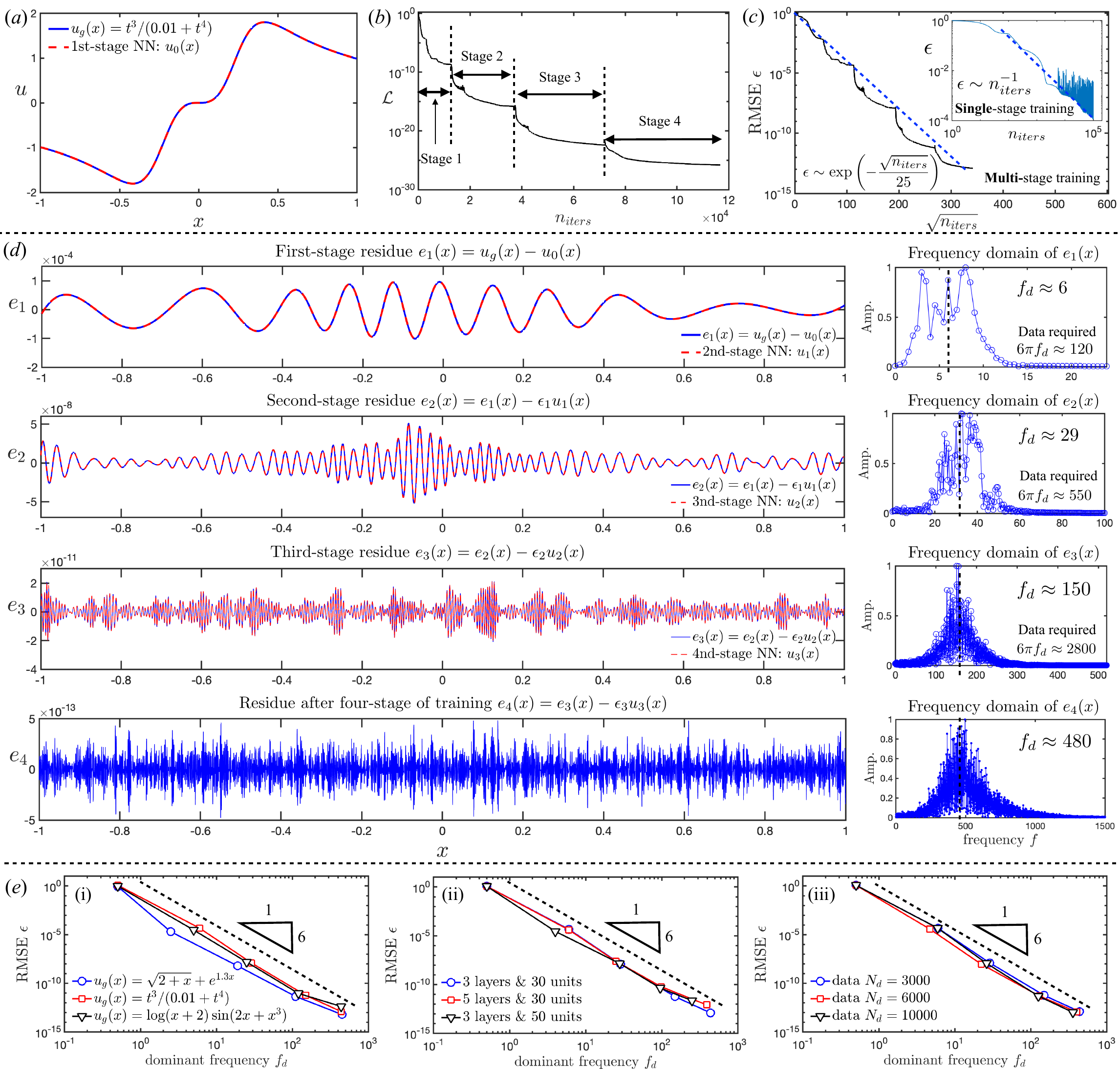} \vspace{-0.5em}
	\caption{\label{fig:RegmNN64}  {\bf Multi-stage neural networks}. ($a$) Fitting of the first-stage neural network (red dashed curve) to the data from a given target function (blue curve). ($b$) Training loss $\mathcal{L}$ over the iterations based on multi-stage training scheme. ($c$) Evolution of the root mean square value $\epsilon$ of the error $e_n(x)$ over the iterations, which follows $\epsilon \sim \exp(-\sqrt{n_{iters}}/25)$, close to an exponential decay. However, for single-stage training ($c$-inset), the error convergence only follows a linear decay, $\epsilon \sim 1/n_{iters}$. ($d$) Fitting of higher-stage networks to the error of lower-stage training. Frequency domain of the error $e_n(x)$ for different stages are shown in the right column. After four stages of training, the error between the data and combined networks is close to the machine precision of a double float (64-bit) ($e$) Relation of the dominant frequency $f_d$ and the root mean square value $\epsilon$ of the error $e_n(x)$ after different stages of training follows a power law \eqref{eq:fde} with an exponent $\alpha$ independent of (i) target functions, (ii) neural network size, (iii) and the number of data points.}
\end{figure}

Following Algorithm \ref{alg:cap}, figure \ref{fig:RegmNN}($b$) shows the result of the second-stage training, which fits the residue $e(x)$ between the first-trained network $u_0(x)$  (figure \ref{fig:RegmNN}$a$) and the original training data $u_g(x)$ from \eqref{eq:eqn1}. In comparison to the single-stage training where the loss plateaued $O(10^{-8})$, the two-stage training dramatically reduced loss to $O(10^{-14})$ (figure \ref{fig:RegmNN}$c$), resulting in the fitting error $e(x)$ between the two-stage trained networks and the data reaches the machine precision $O(10^{-7})$ for a 32-bit single float (figure \ref{fig:RegmNN}$c$). Unless double-precision (64-bit) is employed, further reduction of loss is unattainable. Moreover, we note that the total number of weights used (figure \ref{fig:RegmNN}$c$) in the two-stage neural networks (around $2\times 3\times 20^2 \approx 2400$) is less than that in the single-stage network (around $3 \times 30^2 \approx 2700$) . This underscores that a larger neural network is not inherently advantageous; an appropriate training scheme is more vital and efficient for the reduction of training loss.

In fact, the power of multi-stage training scheme lies not only in boosting the convergence of training, but also fundamentally enabling neural networks to approximate a target function with arbitrary accuracy as required. We now convert the weights, biases and training data from single-float precision to double precision, and create the third and fourth-stage neural networks in accordance with Algorithm \ref{alg:cap}. Figure \ref{fig:RegmNN64}($d$) shows that the error $e_4(x)$ between the sum of four-staged networks and the data successfully approaches the machine precision of a 64-bit double float. As long as higher-precision floating-point is used, the error can be further reduced with additional stages of training.

\begin{figure}
	\centering
	\includegraphics[width=1\textwidth]{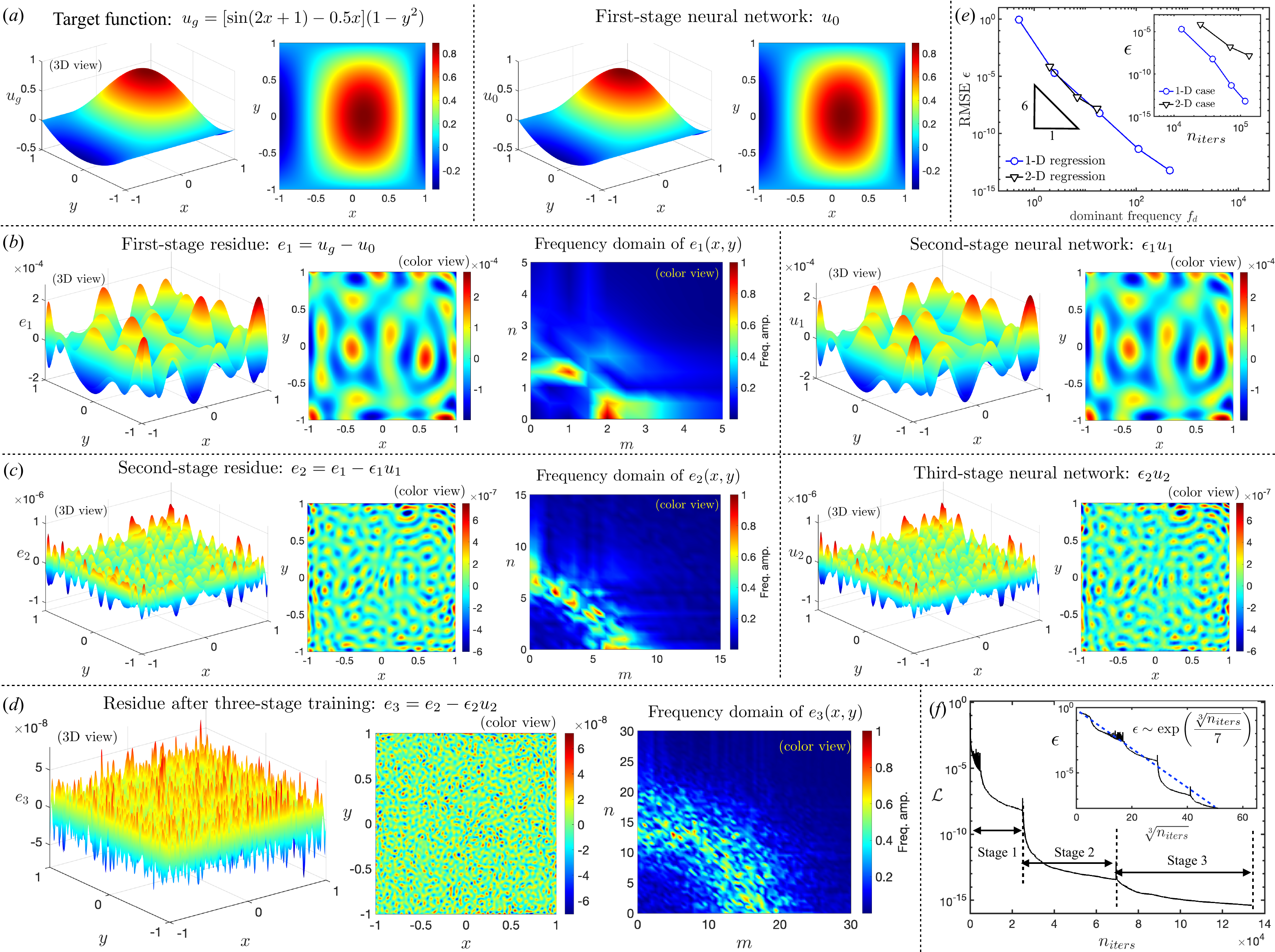}
	\caption{\label{fig:Reg2DmNN} {\bf Multi-stage neural networks for a 2D target function}. ($a$) Fitting of first-stage neural network $u_0(x,y)$ to the data from a 2D target function $u_g(x,y)$. ($b$-$d$) Fitting of higher-stage networks $u_i(x,y)$ to the error $e_i(x,y)$ of lower-stage training. Frequency domain of the error at each stage is given. ($e$) Relation of the dominant frequency $f_d$ with the root mean square value $\epsilon$ of the error $e_n(x,y)$ after different stages of training follows the same power law with the 1D problem, of which the exponent $\alpha = 1/6$. ($f$) Training loss $\mathcal{L}$ over iterations of the multi-stage neural networks. The inset shows that the evolution of the root mean square error $\epsilon$ over iterations for the 2D regression problem follows $\epsilon \sim \exp(-\sqrt[3]{n_{iters}}/7)$, which is slightly slower than that for the 1D problem (see inset of ($e$). }
\end{figure}

Figure \ref{fig:RegmNN64}($c$) shows that the overall convergence rate of the root mean square value $\epsilon$ of the error $e(x)$ between the network and data using multi-stage training scheme follows $\epsilon \sim \exp(-\sqrt{n_{iters}}/25)$, closely approximating exponential decay. In contrast, the regular single-stage training only exhibits a linear decay, $\epsilon \sim 1/n_{iters}$. That is to say, without considering the risk of being trapped in local minima, it would take at least $O(10^{10})$ iterations for a single-stage training to reach an error of $O(10^{-10})$. With the multi-stage training scheme, it only requires $250^2 \approx 6 \times 10^5$ iterations to reach the same error,  which is \textit{four orders of magnitude faster}.

Moreover, we note that the number of data points required for higher-stage training also needs to increase following the criterion \eqref{eq:nscri}. Figure \ref{fig:RegmNN64}($d$) shows that the dominant frequency of residue after three-stage training can reach $f_d = 150$. This implies that the minimal number of training data to guarantee the success of the fourth-stage training needs to be $N_d > 6\pi f_d = 2830$. Figure \ref{fig:RegmNN64}($c$) shows that the relation between the dominant frequency $f_d$ and magnitude $\epsilon$ of the residue after different stages of training empirically follows a power law
\begin{equation}\label{eq:fde}
	f_d \approx f_0 \epsilon^{-\alpha}, \quad \text{with the exponent} \  \alpha = 1/6 \, ,
\end{equation}
where $f_0$ denotes the dominant frequency of the original training data. In practice, we anticipate a gradual increase in the frequency of the residue $e(x)$ corresponding to the decrease in its magnitude, namely the exponent $\alpha$ should be close to 0. Considering the error between a neural network $u(x)$ and target function $u_g(x)$ of magnitude $\epsilon_0$ and dominant frequency $f_d$, the error between the $m$-th derivative of $\hat{u}(x)$ and  $u_g(x)$ becomes,
\begin{equation}
	\frac{d^m}{dx^m} u(x) - \frac{d^m}{dx^m}u_g(x) =\frac{d^m}{dx^m}e(x) \sim (2\pi f_d)^m \epsilon_0 \sim \epsilon_0 ^{1-\alpha m} \, ,
\end{equation}
where we derive the last expression using \eqref{eq:fde}.  We find that when $1-\alpha m > 0$, even if the magnitude of error $\epsilon_0$ is small, the error at high derivatives $m>1/\alpha$ can still exceed 1. This indicates that the trained neural network with high-frequency error tends to miss the high-derivative ($m>1/\alpha$) information of the target function underlying the training data. Hence, our goal is to achieve a smaller $\alpha$ value during training, which enables the neural network to learn the high-derivative information from the data more accurately.

However, figure \ref{fig:RegmNN64}($e$) shows that, for regression problems, the exponent $\alpha$ from multi-stage training scheme appears to be universal, independent of both target functions and neural network settings. To further reduce $\alpha$, high-derivative information about the target function would be required for the training. However, this information is often absent in regression problems, while it is readily available for physics-informed neural networks. The methodology of reducing the exponent $\alpha$ for PINNs will be addressed in a later section (\S\ref{sec:mNN_PINN}). 

Figure \ref{fig:Reg2DmNN}($a$-$d$) shows that the multi-stage training scheme is equally applicable for 2D regression problems. The convergence rate of the loss function for the 2D problem roughly follows  $\epsilon \sim \exp(-\sqrt[3]{n_{iters}}/7)$ (figure \ref{fig:Reg2DmNN}$f$), slower than that for the 1D problem, but still much faster than the linear decay seen with regular single-stage training. Figure \ref{fig:Reg2DmNN}($e$) shows that the relation between the dominant frequency $f_d$ and the root mean square value $\epsilon$ of the 2D residue $e(x,y)$ follows the same power law \eqref{eq:fde} with the exponent $\alpha \approx 1/6$. 

\section{Multi-stage training for physics-informed neural network}\label{sec:PINN}
The multi-stage training scheme is particularly critical when we use neural network to approximate solutions governed by equations, where the demand for precision is high and essential for the usefulness of the solution. Here we apply the multistage idea to the physics-informed neural networks (PINNs) to improve their accuracy to machine precision. Unlike classical numerical method (i.e. finite difference) which can steadily enhance the accuracy of solution by reducing the grid size, PINNs cannot efficiently reduce solution errors merely by adding more collocation points or enlarging the neural network size, similar to the issue seen with regression problems (see \ref{sec:AppA}). This has made PINNs a less favored method for many scientific research that demands high-precision prediction. In this section, we show that the multi-stage training scheme can be extended to address this limitation of PINNs.

The general procedure of multi-stage training scheme for physics-informed neural networks (PINNs) mirrors that for regression problems (Algorithm \ref{alg:cap}). However, {\it two} new challenges emerge when applying multi-stage training to PINNs. {\it First}, for regression problems, we can directly determine the magnitude $\epsilon$ and dominant frequency $f_d$ of the target function for each stage of training from the residue of lower-stage training.  However, for PINNs, these two quantities are not readily obtainable because we lack the exact solution required to estimate the error of lower-stage training. 

In addition, the loss function of PINNs involves both data loss and equation loss, defined as
\begin{gather}\label{eq:lossPINN}
	\mathcal{L} = (1-\gamma)\mathcal{L}_d + \gamma\mathcal{L}_e \qquad \text{with} \\ \label{eq:loss}
	\mathcal{L}_d = \frac{1}{N_d} \sum_{i=1}^{N_d} [{u}(x_i) - u_i]^2  \quad \text{and} \quad 
	\mathcal{L}_e = \frac{1}{N_e} \sum_{j=1}^{N_e} [r(x_j, {u}(x_j) )]^2,
\end{gather}
where $N_d$ represents the number of data points, commonly employed as the boundary condition, and $N_e$ is the number of collocation points, which are utilized to examine the equation residue $r(x, u)$ at various positions within the domain. In comparison to regression problems, $\gamma$, known as the equation weight, is the additional hyper-parameter that balances the significance of the two losses during training. How to determine an appropriate value of $\gamma$ for higher stages of training becomes the {\it second} challenge. Using a simple example, we will demonstrate new algorithms to address these challenges and develop a modified multi-stage training scheme for PINNs.

\subsection{First challenge: magnitude and frequency of higher-stage network}\label{sec:PINN_mag}
As discussed in Section \ref{sec:mNN_reg}, the effectiveness of multi-stage training scheme depends largely on the optimal setting of the higher-stage neural networks $u_{n}$, which is based on the magnitude and frequency of the residue $e_{n}$ between the combined lower-stage networks and the ground truth $u_g$.  However, these pieces of information are not directly accessible for PINNs because we don't have the exact solution $u_g(x)$ to the equation that is required to estimate the error $e(x) = u_g - u_0$ of the lower-stage trained networks. Instead, the only information we have is the equation residue $r(x, u_0)$ associated with the trained first-stage network $u_0$. Thus, understanding the relation between the equation residue $r(x, u_0)$ and the error $e(x)$ of the lower-stage networks with the exact solution is crucial for determining the settings for the higher-stage training of PINNs.

\begin{figure}
    \centering
    \includegraphics[width=1\textwidth]{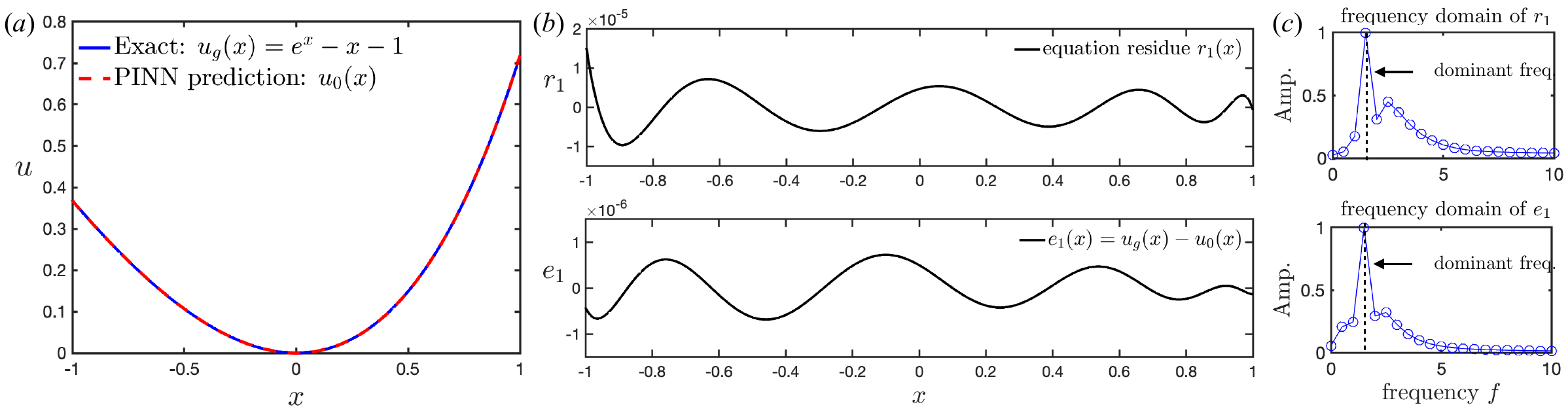}
    \caption{\label{fig:ODE1st} {\bf Comparison of prediction error with equation residue of PINNs}.  ($a$) Exact solution $u_g(x)$ and neural network prediction $u_0(x)$ to equation \eqref{eq:odeqn1}. ($b$)  Comparison of the equation residue $r_1(x, u_0)$ associated with the neural network prediction $u_0(x)$ with the prediction error $e_1(x)$ between $u_0(x)$ and the exact solution $u_g(x)$, which has different magnitude. ($c$) The frequency domain of the equation residue $r_1(x, u_0)$ and the prediction error $e_1(x)$, which has the same dominant frequency.}
\end{figure}

\subsubsection{A simple example}\label{sec:simple}
We consider a first-order ordinary differential equation with the boundary condition
\begin{equation} \label{eq:odeqn1}
\frac{d u}{d x} = u + x \qquad \text{with} \qquad u(0) = 1 \, ,
\end{equation}
which has the exact solution $u_g(x) = e^x - x - 1$. Figure \ref{fig:ODE1st}($a$) shows the single-stage trained network $u_0(x)$ to solve the equation \eqref{eq:odeqn1} via PINN, which matches the exact solution $u_g(x)$ well. The equation residue $r_1(x, u_0)$ associated with the network $u_0(x)$ gives
\begin{equation}\label{eq:resf}
	r_1(x, u_0) = \frac{d u_0}{d x} - (u_0 + x) \, ,
\end{equation}
which has the same dominant frequency with the error $e_1(x) = u_g(x) - u_0(x)$ between the trained network $u_0(x)$ and the exact solution $u_g(x)$. However, the magnitude of equation residue $r_1(x, u_0)$ is one-order of magnitude larger than that of the error $e_1(x)$. To elucidate their relations, we introduce the ansatz,
\begin{eqnarray}\label{eq:ansatz}
	u_g(x) = u_0(x) + \epsilon_1 u_1(x) \qquad \text{with} \quad e_1(x) = \epsilon_1 u_1(x)\, ,
\end{eqnarray}
where $\epsilon_1$ denotes the magnitude of the error $e_1(x)$, and $u_1(x) $ becomes the normalized function within the domain. Substituting the ansatz \eqref{eq:ansatz} into \eqref{eq:odeqn1} and re-arranging the equation gives
\begin{equation} \label{eq:odexpd}
	-\epsilon_1\left( \frac{d u_1}{d x} - u_1 \right) = \frac{d u_0}{d x} - (u_0 + x)\, .
\end{equation}
Recalling \eqref{eq:resf}, the right-hand side of \eqref{eq:odexpd} is the equation residue. Thus, the relation between the prediction error $e_1(x)$ and the equation residue $r_1(x, u_0)$ gives
\begin{equation} \label{eq:err_rel}
	-\epsilon_1 \left( \frac{d u_1}{d x} - u_1 \right)  = r_1(x, u_0)\, ,
\end{equation}
which also becomes the governing equation for the second-stage training and $u_1(x)$ is the second-stage neural network. The boundary condition of $u_1$, based on \eqref{eq:odeqn1} and \eqref{eq:ansatz}, is
\begin{equation}
\epsilon_1 u_1(0) = 1 - u_0(0) \qquad \Longrightarrow \qquad u_1(0) = \frac{ 1 - u_0(0)}{\epsilon_1}\, .
\end{equation}
With the appropriate setting of the equation weight $\gamma$ (as discussed in a later section \S\ref{sec:PINN_gamma}), the data loss of the first-stage training should be much smaller than that of the equation loss. This indicates that the error $e_1(x)$, as well as $u_1(x)$, has much smaller value at the boundary than within the domain. Namely, the boundary condition of $u_1(0)$ can be considered as 0. 

With zero boundary conditions, the magnitude and frequency of the solution $u_1(x)$ are governed by the source function. For a linear equation, the dominant frequency of $u_1(x)$ must be equal to that of the source function, namely the equation residue $r_1(x,u_0)$. Otherwise, the equation cannot be balanced in the frequency domain.  

From \eqref{eq:err_rel}, the magnitude $\epsilon_1$ of the error $e_1(x)$ also appears to be the same as that of the equation residue $r_1(x, u_0)$. However, this is only true when the solution $u_1(x)$ is a low-frequency function. For a high-frequency function, its derivative, which represents its local gradient, becomes large and scales as $O(2\pi f_d$), as discussed in Section \ref{sec:mNN_reg}, where $f_d$ is the dominant frequency of the function. Given that $u_1(x)$ shares the same dominant frequency with the equation residue $r_1(x, u_0)$, the magnitude $\epsilon_1$ of the error $e_1(x)$ between the network $u_0(x)$ and exact solution $u_g(x)$ can be determined by equating the magnitudes of the leading-order terms on both sides of the equation \eqref{eq:err_rel}, which gives,
\begin{equation}\label{eq:err_mag}
	2 \pi f_d \epsilon_1 \sim \epsilon_{r_1} \qquad \Longrightarrow \qquad \epsilon_1 = \frac{\epsilon_{r_1}}{2\pi f_d}  \quad \text{with} \quad \epsilon_{r_1} = \text{RMS}(r_1(x,u_0))
\end{equation}
where we use the root mean square (RMS) value $\epsilon_{r_1}$ to represent the magnitude of the equation residue $r_1(x,u_0)$. Figure \ref{fig:ODE1st}($c$) shows that the dominant frequency for the equation residue $r_1(x, u_0)$ and prediction error $e_1(x)$ are the same, around $f_d \approx 1.5$. Based on \eqref{eq:err_mag}, the magnitude $\epsilon_1$ of the error should be $2\pi f_d \approx 10$ times less than that of the equation residue, consistent with the result shown in figure \ref{fig:ODE1st}($b$). 

\subsubsection{Magnitude and frequency estimation for general differential equations}\label{sec:PINNmf}
To extend the relations between the properties of equation residue $r_1(x, u_0)$ and prediction error $e_1(x)$ for general equations, we now consider a general form of ordinary differential equations
\begin{equation}\label{eq:odegen}
	\mathcal{N}\left(x, u, u^{(1)}, ... u^{(m)}\right) = F(x)   \qquad \text{with} \ \ u^{(i)} = \frac{d^{i}u}{dx^{i}} \ \text{ for } \ i = 1, 2, ..., m
\end{equation}
where $\mathcal{N}$ is a nonlinear differential operator that involves $x$, $u$ and its derivative $u^{(i)}$ at different orders. $m$ represents the highest order of derivative of $u$ in the equation. $F(x)$ is a source function with known expression.  We denote $u_g(x)$ as the exact solution to the equation and $u_0$ the first-stage neural network prediction. By introducing the ansatz \eqref{eq:ansatz} and substituting into \eqref{eq:odegen}, we have 
\begin{equation}\label{eq:odegen1}
	\mathcal{N}\left(x, (u_0 + \epsilon_1 u_1), [u_0 + \epsilon_1 u_1]^{(1)}, ..., [u_0 + \epsilon_1u_1]^{(m)}\right) = F(x)  
\end{equation}
Considering that the first-stage neural network $u_0$ captures the main variation of the exact solution $u_g$, the magnitude $\epsilon_1$ of the error $e_1(x)$ between $u_g$ and $u_0$ would then be much smaller than one, namely $\epsilon_1 \ll 1$. In that case, the equation \eqref{eq:odegen1} can be rewritten in terms of a Taylor expansion of the nonlinear function $\mathcal{N}$. After re-arrangement, it gives
\begin{equation}\label{eq:odegen2}
	-\epsilon_1 \left(  \left. \frac{\partial \mathcal{N}}{\partial u} \right|_{u = u_0} u_1 +  \left. \frac{\partial \mathcal{N}}{\partial u^{(1)}} \right|_{u = u_0} u_1^{(1)} + ... +  \left. \frac{\partial \mathcal{N}}{\partial u^{(m)}} \right|_{u = u_0}   u_1^{(m)}\right) + O(\epsilon_1^2) = \mathcal{N}\left(x, ..., u_0^{(m)}\right) - F(x)  
\end{equation}
where $u$ and its derivative $u^{(i)}$ at different orders are considered as separate independent variables of the function $\mathcal{N}$. Because $\epsilon_1 \ll 1$, all the nonlinear terms of $u_1$ fall into the high-order $O(\epsilon_1^2)$ term, and can generally be disregarded. This suggests that regardless of whether the original equation is linear or nonlinear, the governing equations for higher-stage networks essentially become {\it linear} equations. This is a key factor that ensures the success of multi-stage training scheme for PINNs.  

Since the right-hand side of \eqref{eq:odegen2} is the equation residue of $u_0$, the final equation for $u_1$ gives
\begin{equation}\label{eq:odegen3}
	\epsilon_1 \left(  \left. \frac{\partial \mathcal{N}}{\partial u} \right|_{u = u_0} u_1 +  \left. \frac{\partial \mathcal{N}}{\partial u^{(1)}} \right|_{u^{(1)} = u^{(1)}_0} u_1^{(1)} + ... +  \left. \frac{\partial \mathcal{N}}{\partial u^{(m)}} \right|_{u^{(m)} = u^{(m)}_0}   u_1^{(m)}\right) = r_1(x, u_0)
\end{equation}
or, in a short form,
\begin{equation}\label{eq:odegen32}
	-\epsilon_1 \sum_{k=0}^{m} \beta_k u_1^{(k)} = r_1(x, u_0)  \qquad \text{with} \quad \beta_k = \left. \frac{\partial \mathcal{N}}{\partial u^{(k)}} \right|_{u = u_0} \quad \text{and}\quad  u_1^{(k)} = \frac{d^k u_1}{dx^k}
\end{equation}
As mentioned earlier, if $u_0$ is correctly trained, the boundary condition for $u_1$ should be close to 0. Given that \eqref{eq:odegen32} is linear, the magnitude and frequency of $u_1$ should be determined from the equation residue $r_1(x, u_0)$ by matching the magnitude and frequency of the dominant term (the term with the largest magnitude) on the left-hand side of \eqref{eq:odegen32} with that of $r_1(x, u_0)$. 

Considering a physical equation with coefficients of similar scale before each term , and assuming $u_1$ to be a high-frequency function with a dominant frequency far exceeding that of $u_0$, the dominant term on the left-hand side of \eqref{eq:odegen32} is expected to be the one involving the highest-order derivative of $u_1$, namely $\epsilon_1 \beta_m u_1^{(m)}$.   We denote the dominant frequency of $\beta_m$  and $r_1(x, u_0)$ as $f_d^{(\beta)}$ and $f_d^{(r)}$, respectively. Then, the dominant frequency $f_d^{(1)}$ of $u_1$ gives
\begin{equation}\label{eq:fd0}
	f_d^{(\beta)} + f_d^{(1)} = f_d^{(r)}  \qquad \Longrightarrow \qquad  f_d^{(1)} = f_d^{(r)} - f_d^{(\beta)}.
\end{equation}
Given that $\beta_m$ is a function only with respect to the lower-order network $u_0$, and the dominant frequency $f_d^{(1)}$ of the higher-stage network $u_1$ is much larger than that of $u_0$. Thus, even if $\beta_m$ is a highly-nonlinear function, we have $f_d^{(1)} \gg f_d^{(\beta)}$. Combined with \eqref{eq:fd0}, we have
\begin{equation}\label{eq:fd}
	f_d^{(1)} = f_d^{(r)} - f_d^{(\beta)} \approx f_d^{(r)} 
\end{equation}
where the dominant frequency of $u_1$ is mainly governed by that of equation residue. With the dominant frequency determined, the magnitude $\epsilon_1$ of the error between $u_g$ and $u_0$ can be derived by matching the magnitude of the term $\epsilon_1\beta_m u_1^{(m)}$ and $r_1(x, u_0)$, which gives
\begin{gather}\label{eq:mag}
	 \epsilon_1 \cdot \epsilon_\beta \left[ 2\pi f_d^{(r)} \right]^m = \epsilon_{r_1}  \qquad \Longrightarrow \qquad  \epsilon_1 = \frac{\epsilon_{r_1}}{\left[2\pi f_d^{(r)}\right]^m \epsilon_\beta} \\ \notag
	\text{with} \qquad \epsilon_\beta = \mathrm{RMS}(\beta_m)    \quad  \text{and} \quad \epsilon_{r_1} = \mathrm{RMS}(r_1(x, u_0))
\end{gather}
where we use the root mean square (RMS) value $\epsilon_\beta$ and $\epsilon_{r_1}$ to represent the magnitude of $\beta_m$ and equation residue $r_1(x, u_0)$, respectively. 

The relations \eqref{eq:fd} and \eqref{eq:mag} can also be generalized to {\it  partial} differential equations, for which we need to calculate the dominant frequency of $u_1$ with respect to each independent variable $x_i$, namely
\begin{equation}\label{eq:fd2}
	 f_d^{(1, x_i)} \approx f_d^{(r, x_i)} \qquad \text{for} \quad i = 1, 2, ..., N
\end{equation}
where $N$ is the total number of independent variables of the equation. Then, the magnitude $\epsilon_1$ of the error between the first-stage network $u_0$ and the exact solution $u_g$ would be
\begin{gather}\label{eq:mag2}
    \epsilon_1 = \frac{\epsilon_{r_1}}{\epsilon_\beta \left[2\pi f_d^{(r, x_1)}\right]^{m_1} \cdot \left[2\pi f_d^{(r, x_2)}\right]^{m_2} ... \left[2\pi f_d^{(r, x_N)}\right]^{m_N}} 
\end{gather}
where $m_1 + m_2 + ... + m_N$ represents the highest order of partial derivative of $u_1$ in the equation. For most equations, the relations \eqref{eq:fd}-\eqref{eq:mag2} are sufficiently accurate to estimate the magnitude and frequency of the network for higher-stage PINN training. However, there are two types of nonlinear equations where these relations may not hold exactly. They are discussed in \ref{sec:AppC}. 

\begin{figure}
	\centering
	\includegraphics[width=1\textwidth]{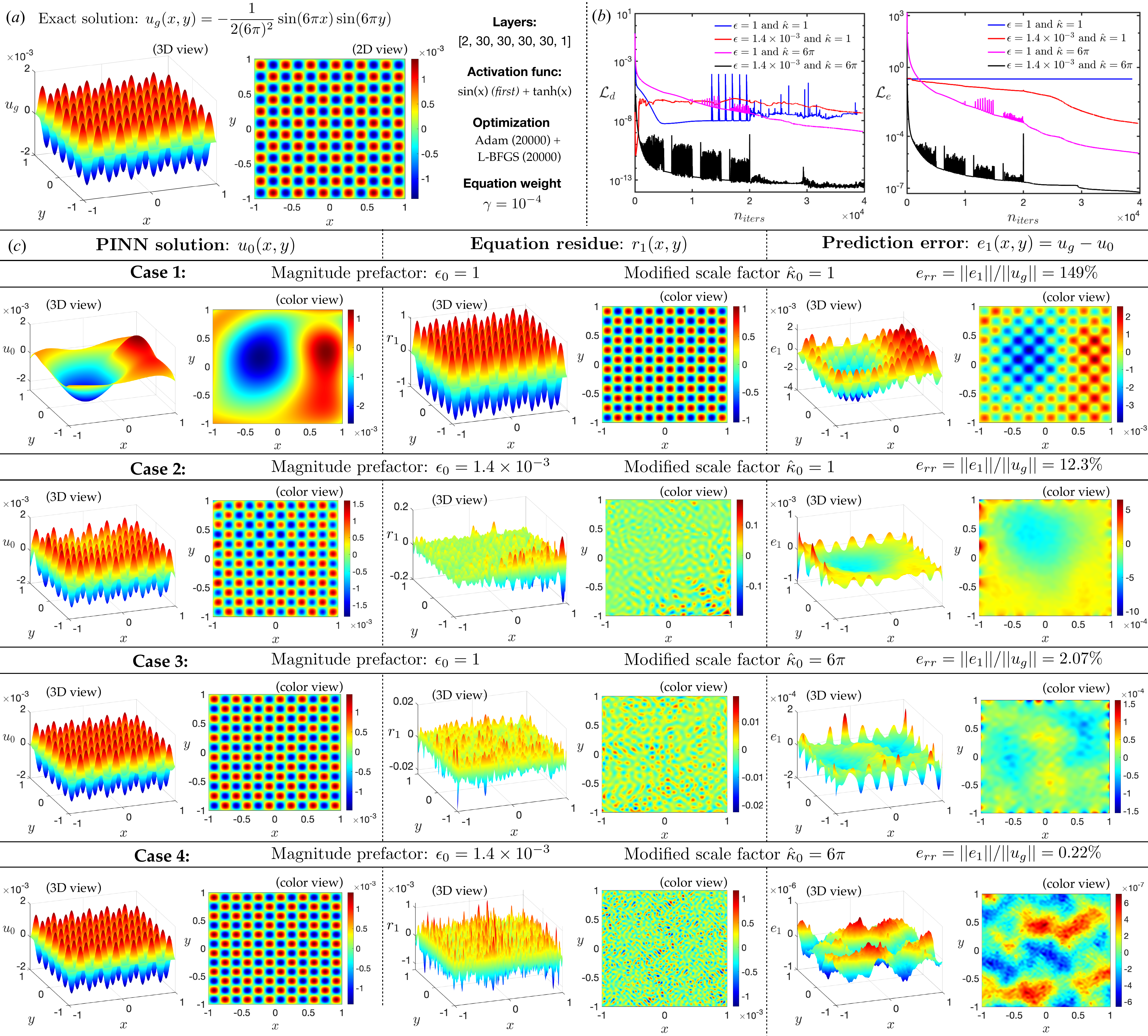} 	\vspace{-0.2em}
	\caption{\label{fig:PHFsin3} {\bf Importance of rescaling PINN magnitude and frequency}. ($a$) Exact solution $u_g(x,y)$ to equation \eqref{eq:Pcase3} and the general setting of PINNs for 4 cases. ($b$) Evolution of the data loss and equation loss over the training of solving \eqref{eq:Pcase3} via PINNs for different magnitude prefactor $\epsilon_0$ and modified scale factors $\hat{\kappa}_0$. ($c$) Trained network $u_0(x,y)$ of the solution to \eqref{eq:Pcase3}, the associated equation residue $r_1(x,y, u_0)$ and prediction error $e_1(x,y)$ under different magnitude prefactor $\epsilon_0$ and modified scale factor $\hat{\kappa}_0$ (Case 1-4). The trained network with $\epsilon_0$ and $\hat{\kappa}_0$ from \eqref{eq:fd2} and \eqref{eq:mag2} gives the prediction with the lowest relative error $e_{rr}$. } \vspace{-0.6em}
\end{figure}

\subsubsection{Importance of magnitude and frequency for higher-stage PINN training }
The proper setting of the magnitude and frequency of a neural network, as shown to be essential for regression problems in Section \ref{sec:mNN_reg}, is critical for physics-informed neural networks, especially during higher-stage training.  To illustrate this, we use a Poisson equation with a high-frequency source function to represent the equation residue, and zero boundary conditions. This setup effectively mimics the governing equation for higher-stage PINN training. The equation reads
\begin{equation}\label{eq:Pcase3}
	u_{xx} + u_{yy} = -\sin(6\pi x)\sin(6\pi y) \qquad \text{with} \quad  u(x,\pm 1) = u(\pm 1, y) = 0 
\end{equation}
At first glance, one might assume that the solution $u$ has the same order of magnitude with the source function, which is $O(1)$. However, based on the analysis \eqref{eq:fd2} in Section \ref{sec:PINNmf}, the solution should have a dominant frequency $f_d = 3$ with respect to both independent variables $x$ and $y$. Given that the highest order of derivative in \eqref{eq:Pcase3} is $m = 2$,  the magnitude of the solution $u$ can, then, be derived from \eqref{eq:mag2}, as $1/[2 \pi f_d]^2 \sim O(10^{-3})$, as shown in the exact solution,
\begin{equation}
	u_g(x,y) = \frac{1}{2(6 \pi)^2}\sin(6\pi x)\sin(6\pi y)\, .
\end{equation}
 Figure \ref{fig:PHFsin3} shows the neural network predicted solution $u_0$ to \eqref{eq:Pcase3} via PINNs under different setting of magnitude (via magnitude prefactor $\epsilon_0$) and frequency (via modified scale factor $\hat{\kappa}_0$). In these cases, we assume that the correct value of the equation weight $\gamma$ is used. As shown in figure \ref{fig:PHFsin3}($b$), only when both magnitude and frequency are correctly set in accordance with \eqref{eq:fd2} and \eqref{eq:mag2} does the neural network successfully converge to the exact solution at a rapid convergence rate. 

\subsubsection{Algorithm for determining the solution magnitude for higher-stage PINN training }
Besides the theoretical relations \eqref{eq:fd}-\eqref{eq:mag2} derived in Section \ref{sec:PINNmf}, we also develop a general algorithm to determine the magnitude of the solution to linear differential equations with high-frequency source functions and zero boundary conditions to mimic the higher-stage training of PINN. The algorithm can subsequently be combined with Algorithm \ref{alg:cap} to extend the multi-stage training scheme for PINNs. The specific steps of the algorithm are given in Algorithm \ref{alg:epsil}.

\begin{algorithm}[h]
	\caption{Determine the magnitude $\epsilon$ of solution to a linear equation with {\it high-frequency} source function and {\bf zero} boundary condition }\label{alg:epsil}
	\begin{algorithmic}[1]
		\item Write a linear equation in terms of the source function $s({\bf x})$ (function without dependent variable) as
        \begin{equation}
			\mathcal{N} [{\bf x}, u({\bf x})]  = s({\bf x}).
		\end{equation}
        where $\mathcal{N}[\:\cdot\:]$ indicates the differential operator involved in a given equation and ${\bf x} = (x_1, x_2, ...)$ represents the independent variables. 
        \vspace{0.5em}
		\item Define a guess solution based on source function $s({\bf x})$ as
		\begin{equation}\label{eq:ugess}
			u_s({\bf x},\alpha)  = \alpha s({\bf x}).
		\end{equation}
        where the coefficient $\alpha$ can be initially set to be 1.  \vspace{0.5em}
		\item Substitute the guess solution $u_s$ \eqref{eq:ugess} and calculate the associated differential operator $\mathcal{N}_s$
		\begin{equation}
			\mathcal{N}_s({\bf x},  \alpha) = \mathcal{N}[{\bf x}, u_s({\bf x}, \alpha)]
		\end{equation}
        In practice, for complicated equations, the differential operator $\mathcal{N}_s$ can be more easily calculated by adding the equation residue $r_s({\bf x}, u_s)$ associated with the guess solution with the source term $s({\bf x})$, namely 
        \begin{equation}
			\mathcal{N}_s({\bf x},  \alpha) = r_s[{\bf x}, u_s({\bf x})] + s({\bf x})
		\end{equation}
        
		\item Introduce an iteration process with learning rate $\eta$. Define the criterion for the coefficient $\alpha$
		\begin{equation}\label{eq:Rcrit}
			0.1 < R(\alpha) = \frac{||N_s({\bf x},\alpha)||}{||s({{\bf x}})||} < 10
		\end{equation}
		where $||\cdot||$ represents the $l_2$-norm.  If the ratio $R$ falls outside this specified range, update $\alpha$ at each iteration by
		\begin{equation}\label{eq:recur}
			\alpha_{n+1} \ = \alpha_n  \left[\frac{1}{R(\alpha_n)}\right]^\eta = \alpha_n \left(\frac{||s({\bf x})||}{||\mathcal{N}_s({\bf x},\alpha_n)||}\right)^\eta
		\end{equation}
		and re-calculate $R(\alpha)$ based on the updated value of $\alpha$ until $R$ meets the criterion \eqref{eq:Rcrit}. \vspace{0.5em}
		\item Finally, the solution magnitude $\epsilon_0$ can be estimated by multiplying $\alpha$ from Step 4 with the root mean square value $\epsilon_s$ of the source function, namely
        \begin{equation}\label{eq:mags}
            \epsilon_0 = \alpha \cdot \epsilon_s \qquad \text{where} \quad  \epsilon_s = \mathrm{RMS}(s(x,y))
        \end{equation}
    
	\end{algorithmic}
\end{algorithm}

The principle underlying the algorithm is based on the fact that the dominant frequency of the solution $u({\bf x})$ mirrors that of the source function $s({\bf x})$. Therefore, the amplification effect of the derivative, attributable to the high-frequency property of the solution, can be well-estimated by taking $s({\bf x})$ as the guess solution. We define the ratio $R$ as the magnitude of the differential operator $\mathcal{N}_s$ relative to that of the source function $s({\bf x})$.  

If $R$ is larger than 10, the magnitude of the differential operator $\mathcal{N}_s$ associated with the guess solution \eqref{eq:ugess} much larger than the source function $s({\bf x})$. In that case, the magnitude of the guess solution should be reduced by decreasing $\alpha$. Conversely, when $R$ is less than 0.1, it suggests that differential operator $\mathcal{N}_s$ associated with the guess solution \eqref{eq:ugess} is too small. Hence, we should increase the magnitude of the solution by increasing $\alpha$. The recursive relation \eqref{eq:recur} in Algorithm \ref{alg:epsil} is designed to achieve this objective. Here, the learning rate $\eta$ is a user-defined positive hyper-parameter, which determines the rate at which $\alpha$ and $R(\alpha)$ converges to satisfy the criterion \eqref{eq:Rcrit}. Finally, the magnitude of the solution $\epsilon$ can be estimated using \eqref{eq:mags}.

Applying Algorithm \ref{alg:epsil} to equation \eqref{eq:Pcase3}, we obtain that $\epsilon = 1.41\times 10^{-3}$, which is very close to the magnitude of the exact solution, $\epsilon = 1/[2(6\pi)^2] = 1.43 \times 10^{-3}$.  Here, we note that Algorithm \eqref{alg:epsil} is mainly applicable to linear differential equations with a single dependent variable. For nonlinear equations or a group of differential equations with multiple dependent variables, a more advanced algorithm may be required to determine the magnitude for each variable, which is beyond the scope of this paper.

\subsection{Second challenge: equation weight $\gamma$ for higher-stage network} \label{sec:PINN_gamma}
Equation weight $\gamma$, as shown in \eqref{eq:lossPINN}, is a hyper-parameter to balance the contribution of data loss and equation loss in the loss function for physics-informed neural networks (PINNs).  In the context of PINNs as a differential equation solver, boundary conditions are often implemented as data loss and the governing equations constitute the equation loss. Given that boundary conditions determine the uniqueness of the solution, a general rule of thumb is weighting the data loss more than the equation loss in the loss function \cite{mcclenny2020self}. This ensures that the boundary conditions are prioritized and satisfied during the training. 

\begin{figure}
	\centering
	\includegraphics[width=\textwidth]{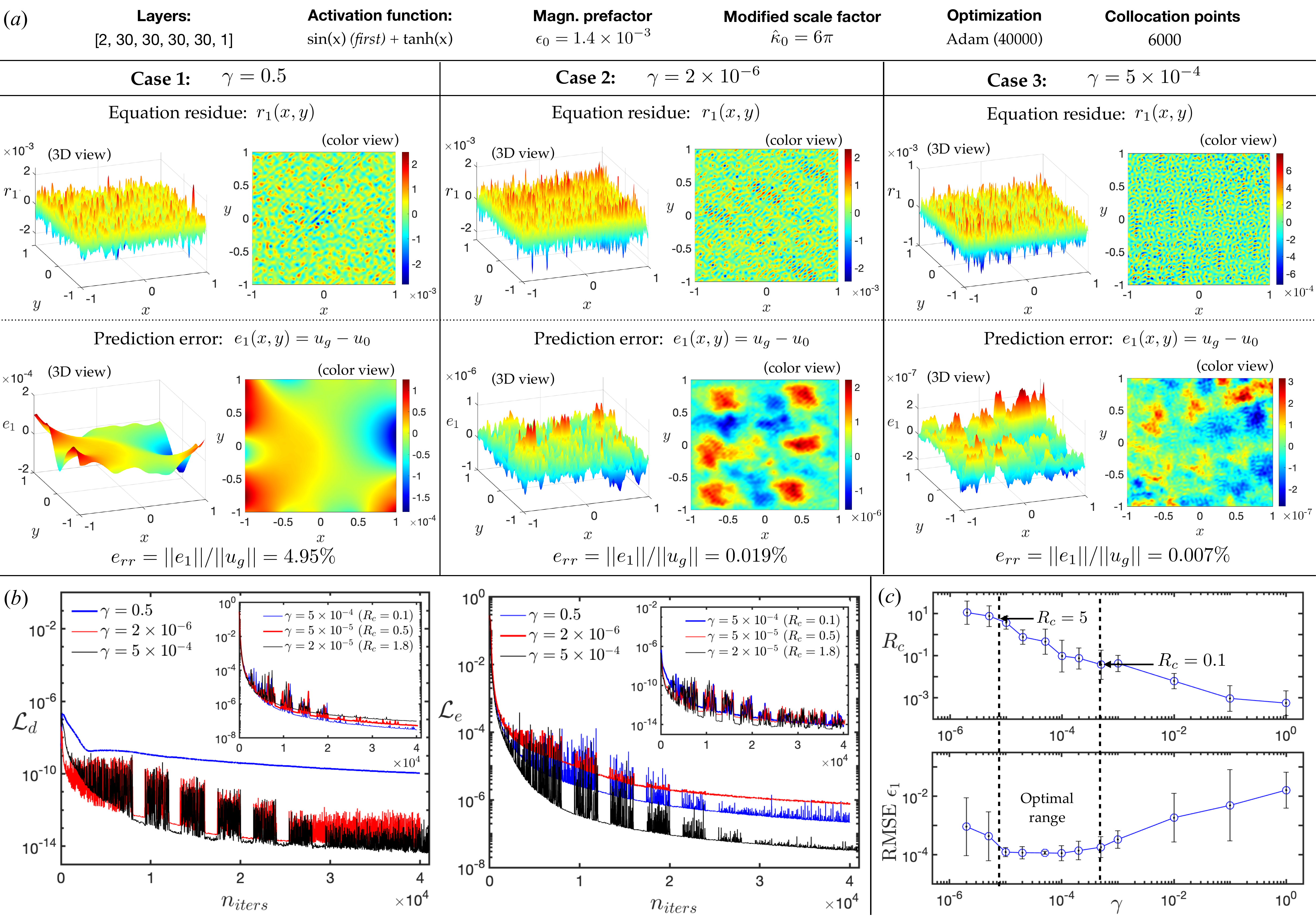}
	\caption{\label{fig:PHF} {\bf Importance of the equation weight $\gamma$ of PINNs}. ($a$) Equation residue $r_1(x,y)$ and prediction error $e_1(x,y)$ of solving \eqref{eq:Pcase3} via PINN for different $\gamma$.   ($b$) Evolution of data loss and equation loss over the training for different $\gamma$ shown in ($a$). The inset shows the loss evolution for different $\gamma$ that satisfy the criterion \eqref{eq:recur2}, which are close to each other. ($c$) The root mean square value $\epsilon_1$ of the prediction error $e_1(x,y)$ (lower panel), and the corresponding ratio $R_c$ of the data loss convergence rate over that of equation loss (upper panel) as a function of $\gamma$. The optimal range of $\gamma$ with minimal prediction error corresponds to $0.1<R_c<5$. Error bars show the standard deviation of five repetitive experiments with different random initialization. }
\end{figure}

The relative contribution of data loss and equation loss in the loss function \eqref{eq:lossPINN} is the ratio of the first term, $I_1 = (1-\gamma) \mathcal{L}_d$, to the second term, $I_2 = \gamma \mathcal{L}_e$ in \eqref{eq:lossPINN}, i.e. $I_1/I_2$.  For normalized linear differential equations that have low-frequency solutions, such as \eqref{eq:odeqn1}, the equation loss, $\mathcal{L}_e$, remains the same order of magnitude with the data loss $\mathcal{L}_d$, around $O(1)$.  In that case, by setting $0.1<\gamma<0.5$, we can ensure that the contribution of data loss $I_1>I_2$ in the loss function \eqref{eq:lossPINN}.  However, this setting of $\gamma$ does not hold for differential equations with high-frequency solutions.  A systematic mathematical justification was provided by a prior study \cite{van2022optimally}, showing that the magnitude of the equation loss increases with the frequency of the solution. When considering the same Poisson equation \eqref{eq:Pcase3} with zero boundary conditions and a high-frequency source function, similar to the equation we solve during higher-stage training, the magnitude of equation loss at the beginning of the training can be estimated as
\begin{eqnarray}\label{eq:P1eqnL}
	\mathcal{L}_e = \frac{1}{N_e}\sum_{i=0}^{N_e}  \left[u_{xx} + u_{yy} + \sin(6\pi x_i)\sin(6\pi y_i) \right]^2  \sim  O(1),
\end{eqnarray}
which is determined by the magnitude of the source function $\sin(6\pi x)\sin(6\pi y)$. The magnitude of data loss at the beginning of training reads
\begin{equation}\label{eq:P1dataL}
	\mathcal{L}_d =  \frac{1}{N_1}\sum_{i=0}^{N_1} [u(x_i, \pm 1)]^2  + \frac{1}{N_2}\sum_{j=0}^{N_2}  [u(\pm 1, y_j)]^2 \sim O(u^2)
\end{equation}
which is determined by the initial magnitude of the solution. Considering that the magnitude $\epsilon_0$ of the solution has been estimated from the relation \eqref{eq:mag2} or Algorithm \ref{alg:epsil}, which gives $\epsilon_0 \approx 1/(6\pi)^2$ for the solution to \eqref{eq:Pcase3}, the magnitude of the data loss thus becomes
\begin{equation}\label{eq:P1dataL2}
	\mathcal{L}_d  \sim O(\epsilon_0^2) \sim O(10^{-6})
\end{equation}
which is six orders of magnitude smaller than the equation loss. If we still use $\gamma \sim O(0.1)$, as $I_2\gg I_1$, the optimization process will primarily focus on minimizing the $I_2$ during the training, largely neglecting the contribution from the data loss.  Utilizing the appropriate values of the magnitude prefactor $\epsilon_0$ and the modified scale factor $\hat{\kappa}_0$ from Section \ref{sec:PINN},  figure \ref{fig:PHF}($b$) shows the evolution of the data loss and equation loss over iterations by setting $\gamma = 0.5$. Compared with the equation loss, which was reduced by seven orders of magnitude in total, the data loss decays at a much slower rate, significantly limiting the errors of the trained neural network \ref{fig:PHF}($c$) to be round 5\%.  

\subsubsection{Theoretical approach of determining $\gamma$ } \label{sec:estlamb}
To improve the accuracy of trained network via PINN, we need to minimize the data loss prior to the equation loss. Therefore, the optimal value of $\gamma$ should yield a larger contribution from the data loss larger than from the equation loss, namely,
 \begin{equation}\label{eq:lamb_c}
 	I_1  = (1-\gamma)\mathcal{L}_d \geq  \gamma \mathcal{L}_e = I_2 \qquad \Longrightarrow \qquad \gamma \leq \frac{\mathcal{L}_d}{\mathcal{L}_e + \mathcal{L}_d},
 \end{equation}
which is consistent with the expression proposed in a prior study \cite{van2022optimally}. For equation \eqref{eq:Pcase3}, with the magnitude of the equation loss and data loss determined from \eqref{eq:P1eqnL} and \eqref{eq:P1dataL} respectively,  equation \eqref{eq:lamb_c} yields $\gamma = 2\times 10^{-6}$. With this $\gamma$,  figure \ref{fig:PHF}($b$) shows that both data loss and equation loss rapidly decrease over the training by more than five orders of magnitude. This suggests that both the equation and boundary condition are progressively satisfied by the network throughout the training.  Although the reduction in equation loss is slightly less than in the case with $\gamma = 0.5$, the relative error $e_{rr}$ between the trained neural network and the exact solution is reduced by more than one hundred times (figure \ref{fig:PHF}$a$). However, generally without prior knowledge of the solution we do not know its corresponding $\mathcal{L}_d$ and $\mathcal{L}_e$, so estimating $\gamma$ theoretically is difficult. Therefore below we develop an alternative approach.

\subsubsection{Algorithm for determining $\gamma$ for general equations} \label{sec:estlamb2}

Besides the theoretical expression \eqref{eq:lamb_c}, we also develop a more general algorithm to determine $\gamma$ through a pre-training process. This approach provides higher accuracy and adaptability for a broad range of problems. As mentioned previously, the optimal value of $\gamma$ should result in similar convergence rates for the data loss and equation loss over the course of training.  We propose a heuristic approach for determining the optimal $\gamma$, which is outlined in Algorithm \ref{alg:lambda}. 

\begin{algorithm}[h]
	\caption{Determine the optimal equation weight $\gamma$ for general PINN training }\label{alg:lambda}
	\begin{algorithmic}[1]
	\item[] {\bf Prerequisite}: normalize the equation with the magnitude of the largest term around $O(1)$ and apply the scale factor and the magnitude prefactor estimated from Algorithm \ref{alg:epsil} to the neural network.  \vspace{0.5em}
	\item Set the initial value of $\gamma$ and calculate the initial data loss and equation loss
	\begin{equation}
		\mathcal{L}_{d}^{(0)} = \mathcal{L}_d(i=0) \qquad \text{and} \qquad \mathcal{L}_{e}^{(0)} = \mathcal{L}_e(i=0) 
	\end{equation}
	where $\mathcal{L}_d^{(i)}$ and $\mathcal{L}_e^{(i)}$ represent the data and equation loss at the $i$-th iteration, respectively. \vspace{0.5em}
	\item Pre-train the neural network for $N_0$ iterations and calculate the minimal data loss and equation loss within the last $N_1 = 0.1N_0$ iterations as
	\begin{equation}
		\mathcal{L}_{d}^{(m)} = \min_{i=0}^{N_1-1} \mathcal{L}_d(N_0-i) \qquad \text{and} \qquad \mathcal{L}_{e}^{(m)} = \min_{i=0}^{N_1-1} \mathcal{L}_e(N_0-i) 
	\end{equation}
	\item Quantify the convergence rate of the data loss and equation loss with 
	\begin{equation}
		C_d =  \frac{\mathcal{L}_{d}^{(0)}}{\mathcal{L}_{d}^{(m)}}  \qquad \text{and} \qquad C_e =  \frac{\mathcal{L}_{e}^{(0)}}{\mathcal{L}_{e}^{(m)}} 
	\end{equation}
	\item Define the ratio of the two convergence as $R_c = C_d/C_e$ and set the criterion 
	\begin{equation} \label{eq:Rcrit2}
		O(0.1) < R_c(\gamma) < O(10)
	\end{equation}
	\item If the ratio is outside the criterion \eqref{eq:Rcrit2} , update $\gamma$ by
	\begin{gather} \label{eq:recur2}
		\gamma \gets \gamma \cdot R_c^\eta
	\end{gather}
	and rerun Step 3 - 6 with the new updated $\gamma$ until the criterion \eqref{eq:Rcrit2} is satisfied.
	\end{algorithmic}
\end{algorithm}

The principle underlying the algorithm lies in estimating the \textit{convergence rates} of both data loss $C_d$ and equation loss $C_e$. This estimation involves calculating the ratio of the initial loss $\mathcal{L}_d^{0}$ and $\mathcal{L}_e^{0}$, to the respective losses $\mathcal{L}_d^{m}$ and $\mathcal{L}_e^{m}$ after a short period of pre-training . Here, we use the minimal value during the last 10\% of the pre-training iterations to calculate $\mathcal{L}_d^{m}$ and $\mathcal{L}_e^{m}$ to counteract any potential spikes in the loss evolution. 

If the convergence rate of the data loss $C_d$ is substantially lower than that of the equation loss $C_e$, it indicates that the $\gamma$ used in training is too large and needs to be reduced. Vice versa. The recursive relation \eqref{eq:recur2} is designed to reach this goal.  $\eta$ can be considered as the learning rate, a hyper-parameter that determines how fast $R_c(\gamma)$ meets the criterion \eqref{eq:Rcrit2}. We note that, when $\gamma$ is updated, one should re-train the neural network from the beginning to compute the updated $R_c(\gamma)$, instead of continuing the previous training. 

Apply Algorithm \ref{alg:lambda} to the equation \eqref{eq:Pcase3} with $N_0$ set to be 500, one gives $\gamma \approx 10^{-4}$. Figure \ref{fig:PHF}($a$) shows the trained network using this $\gamma$, which reaches further higher accuracy than that using $\gamma = 2\times 10^{-6}$ from \eqref{eq:lamb_c}.  Compared with the case of $\gamma=0.5$ and $\gamma = 2\times 10^{-6}$, the convergence rate of both data loss and equation loss when using $\gamma$ from Algorithm \ref{alg:lambda} are maximized (figure \ref{fig:PHF}$b$), leading to the smallest errors of the neural network prediction.

The criterion range in \eqref{eq:Rcrit2} suggests that the training accuracy is not overly sensitive to the value of $\gamma$, provided that the convergence rate of data loss $C_d$ and equation loss $C_d$ remains within the same order of magnitude. Figure \ref{fig:PHF}($c$) shows the optimal range of $\gamma$ that yields the minimal root mean square value $\epsilon_1$ of the error $e_1(x)$ between the trained network and the exact solution to \eqref{eq:Pcase3}. The corresponding range of $R_c$ is found to be $0.1<R_c<5$, aligning with the range in criterion \eqref{eq:recur2}.  The inset of figure \ref{fig:PHF}($b$) shows that the evolution of both data loss $\mathcal{L}_d$ and equation loss $\mathcal{L}_e$ for different $R_c(\gamma)$ with this range \eqref{eq:Rcrit2} are closely matched. For $\gamma=0.5$, the ratio $R_c$, based on Algorithm \ref{alg:lambda}, is found to be $R_c(\gamma=0.5) = 10^{-4}$, which largely deviates from the criterion, thus, resulting in large prediction error.

\subsection{Additional setting of PINN training for higher-stage network} \label{sec:PINN_add}
Besides the most critical settings, i.e. magnitude prefactor $\epsilon$, scale factor $ \kappa$ of frequency, and the equation weight $\gamma$, there are other settings and advanced algorithm developed in the literature that can ensure the success of PINN training for the high-stage networks.

\begin{figure}
	\centering
	\includegraphics[width=\textwidth]{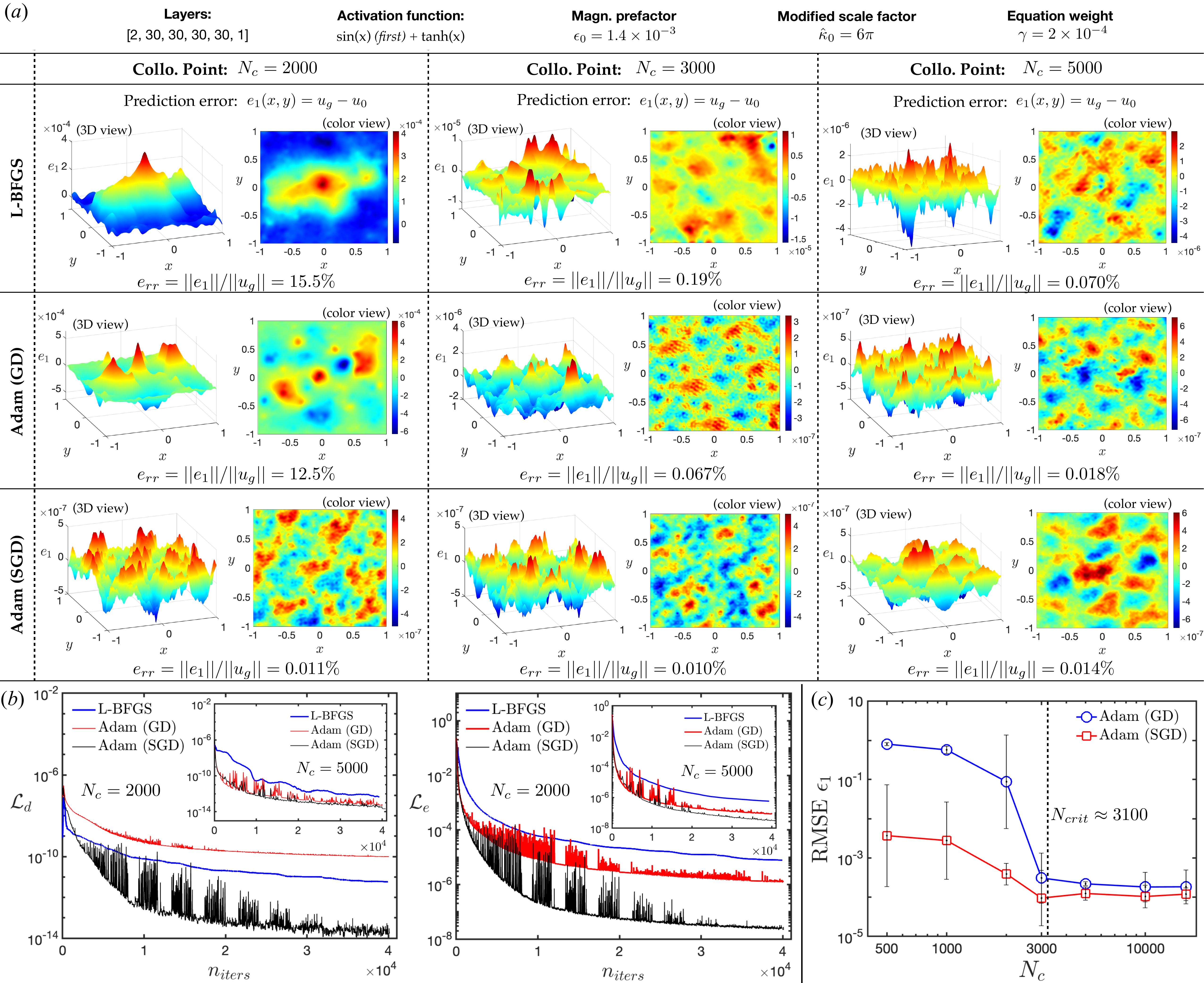}
	\caption{\label{fig:PHF2} {\bf Importance of re-sampling collocation points of PINNs}. ($a$) Comparison of prediction error $e_1(x,y)$ of solving (3.26) via PINNs for different number of collocation points $N_c$ and using different optimizer, including L-BFGS, Adam with gradient descend (GD) (fixed collocation points) and Adam with stochastic gradient descent (re-sample collocation points over the iterations). ($b$) The evolution of data loss and equation loss over the iterations for different optimizer using the number of collocation points $N_c$ below or above (inset) the critical value $N_{crit}$.  ($c$) The relation of the root mean square value $\epsilon_1$ of the prediction error $e_1(x,y)$ with the number of collocation points $N_{c}$ for Adam (GD) and Adam (SGD). When the number of collocation points $N_c$ is less than the critical value $N_{crit} \approx 3100$, stochastic gradient descend can reach better performance for predicting high-frequency solutions. Error bars show the standard deviation of five repetitive experiments with different random initialization.}
\end{figure}

\subsubsection{Optimization method and re-sampling collocation points}\label{sec:opt}
Two other critical settings in the training of high-frequency function includes the selection of optimization method and the number of collocation point. Common choices for PINN training optimizer include Adam and L-BFGS, a second-order quasi Newton method. For general equations with low-frequency solution, L-BFGS is often the preferred optimization method. However, for equations with high-frequency solutions, this is not always the case. Figure \ref{fig:PHF2}($a$) presents a comparison of the loss evolution and final prediction error of trained network between using Adam and L-BFGS for solving the equation \eqref{eq:Pcase3}, where Adam shows a better overall convergence rate. Furthermore, Adam has the added advantage of utilizing stochastic gradient descent (SGD) by re-sampling the collocation points every few iterations \cite{cowen-breen, iwasaki2023}, which shows a even higher convergence rate. 

Collocation points in PINN training is as important as data points in regression problems.  As discussed in Section \ref{sec:freq_reg}, when training the neural network to fit high-frequency data, a sufficient number of data points ($3\pi \approx 10$ per dominant period) are needed to ensure accurate predictions. This principle remains valid for PINN training.  Unlike regression problems, which are limited by the availability of finite data points, PINN could potentially utilize as many collocation points as computationally feasible. For equation \eqref{eq:Pcase3}, the dominant frequency $f_d=3$ in each dimension. Given that the domain is defined in $(x, y) \in [-1, 1]$, there are 6 dominant periods in each dimension. Thus approximately $N_{crit} = (3\pi \times 6)^2 \approx 3100$ collocation points are required.  Figure \ref{fig:PHF2}($a$) compares the accuracy of the trained network for different number of collocation points. For L-BFGS and Adam (GD) with fixed and small number of collocation points, the neural network predictions significantly deviate from the exact solution. When the number of collocation points reach the criterion $N_{crit}$, the prediction error drops sharply and only improves marginally with the addition of more collocation points. 

Compared with using fixed collocation points, predictions using Adam with stochastic gradient descent (SGD) are less sensitive to the number of collocation points. Figure \ref{fig:PHF2}($a$) shows that the prediction error using Adam (SGD) can attain optimal precision even when the number $N_c$ of collocation points falls below the critical value $N_{crit}$. Figure \ref{fig:PHF2}($c$) further compare the root mean square error (RMSE) $\epsilon$ of the trained network between utilizing Adam (GD) and Adam (SGD) for different number of collocation points $N_c$. It confirms that SGD is an essential tool in PINN training for predicting high-frequency solution.

\subsubsection{Advanced methods from the literature: RAR and gPINNs}
Having discussed the essential settings, we note that many advanced algorithms developed in the literature can also largely improve the PINN training of higher-stage networks. Two of most useful methods we found are the adaptive residual-based collocation refinement (RAR) method \cite{lu2021deepxde, qin2022rar} and the gradient-enhanced physics-informed networks (gPINNs) \cite{yu2022gradient} . 

A usual practice in PINN training is to uniformly distribute the collocation points across the domain. However, this approach proves inadequate for equations whose solution feature steep gradients \cite{wu2023comprehensive}. As discussed in Section \ref{sec:mNN_reg}, high-frequency solutions exhibit large gradients throughout the domain. Despite setting a large scale factor to align with the gradient, there remain regions where the local gradient exceeds the averaged gradient $O(2\pi f_d$) with a dominant frequency $f_d$. It can be challenging to minimize the local residue of equation in these areas. To address this issue, we employ the residual-based refinement (RAR \cite{lu2021deepxde}) of collocation point. By continuing adding collocation points in areas of high equation residue throughout the training, the equation residue across the entire domain can be efficiently reduced. This technique thus becomes a vital tool for optimizing PINN training.

An additional method to boost the training performance of PINNs involves incorporating the gradient of the equation residue function $r(x, u)$ into the loss function $\mathcal{L}$, known as the gradient-enhanced physics-informed network (gPINN) \cite{yu2022gradient}. Thus, the loss function can be expressed as,
\begin{gather}\label{eq:lossgPINN}
	\mathcal{L} = (1-\gamma)\mathcal{L}_d + \gamma(\mathcal{L}_e + \gamma_g\mathcal{L}_g)\qquad \text{with} \quad 
	\mathcal{L}_g = \frac{1}{N_g} \sum_{j=1}^{N_g} |\nabla r(x_j, {u}(x_j) )|^2,
\end{gather}
where $N_g$ denotes the number of collocation points used to examine the gradient of the equation residue $r(x, u)$ within the domain. $\gamma_g$ is an additional hyper-parameter, akin to $\gamma$, that control the balance between the equation loss $\mathcal{L}_e$ and gradient loss $\mathcal{L}_g$ during training.

By incorporating the gradient constraint $\mathcal{L}_g$, we obligate the neural networks to learn the high-derivative information of the solution involved in the gradient of the equation. This can significantly improve the convergence rate of the training loss, provided we choose the appropriate value of the weight $\gamma_g$ for the gradient constraint $\mathcal{L}_g$. Analogous to \eqref{eq:lamb_c}, the value of $\gamma_g$ can be estimated by 
 \begin{equation}\label{eq:lamb_g}
	\mathcal{L}_e \geq  \gamma_g \mathcal{L}_g  \qquad \Longrightarrow \qquad \gamma_g \leq \frac{\mathcal{L}_e}{\mathcal{L}_g} \sim \frac{||r||^2}{||\nabla r||^2}
\end{equation}
where $||\cdot||^2$ represents the $l_2$-norm. As discussed in Section \ref{sec:estlamb2}, for the equation with high-frequency solutions,  the equation residue has roughly the same frequency with the solution. Thus, the magnitude ratio of the equation residue $||r||$ with its gradient should scale as $||r||/||\nabla r|| \sim O(2\pi f_d)^{-1}$, where $f_d$ is the dominant frequency of the solution. Thus, the optimal value of $\gamma_g$ can be selected as 
  \begin{equation}\label{eq:lamb_g}
 	\gamma_g = \frac{||r||^2}{||\nabla r||^2} \sim O(2\pi f_d)^{-2}
 \end{equation}
The effect of the gPINN on the higher-stage training is shown and discussed in Section \ref{sec:mNN_PINN}.

\begin{figure}
	\centering
	\includegraphics[width=\textwidth]{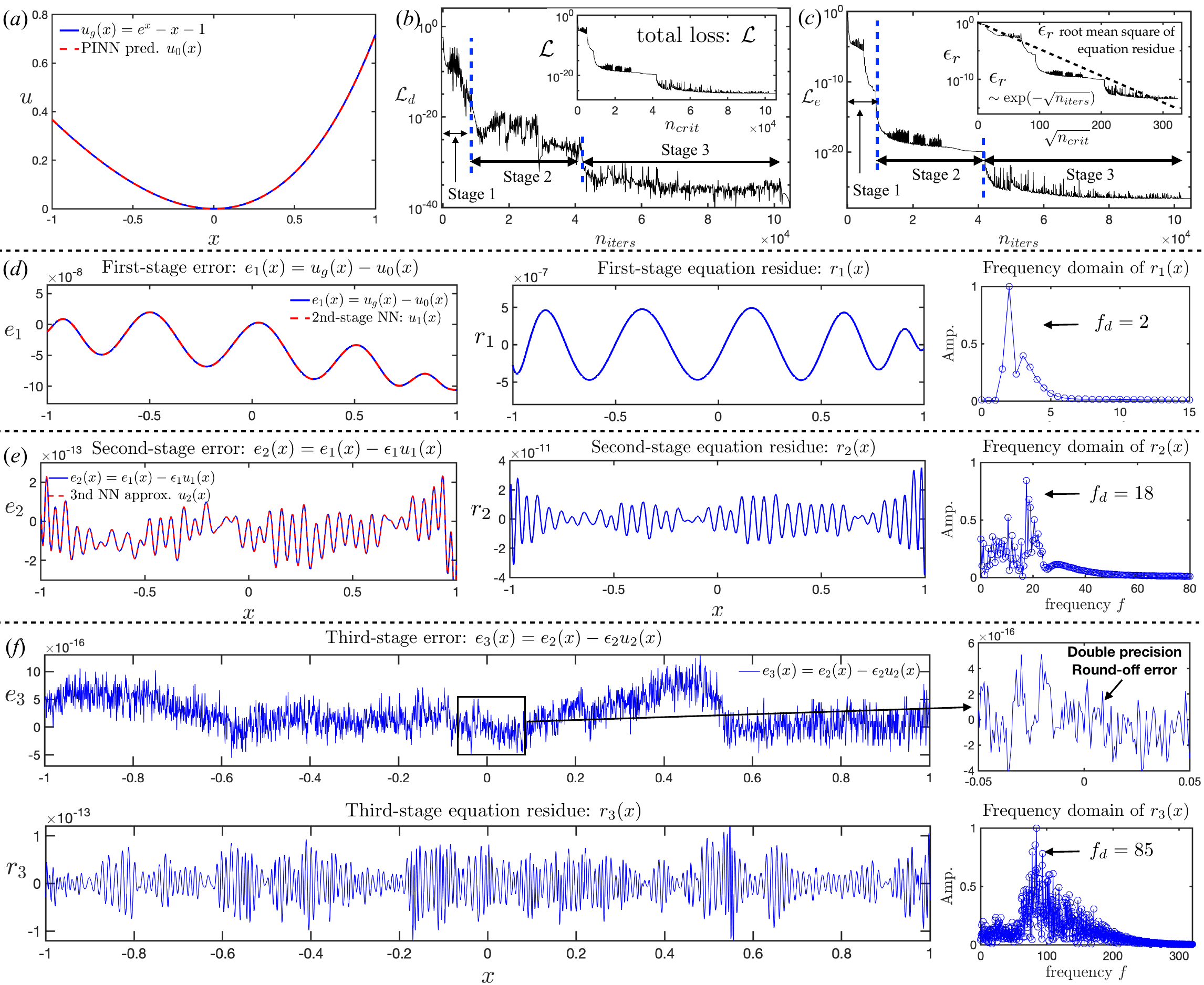}
	\caption{\label{fig:mPINN1d} {\bf Multi-stage gPINNs for 1D equations}. ($a$) Comparison of the first-stage trained neural network (red dashed curve) with the exact solution $u_g(x)$ (blue curve) to equation \eqref{eq:Pcase3}.  ($b$) Data loss and ($c$) equation loss over iterations of three-stage training. The inset of ($b$) shows the evolution of the total loss $\mathcal{L}$ over iterations. The inset of ($c$) shows that the evolution of the root mean square value $\epsilon_r$ of the equation residue $r(x,u)$ of the multi-stage neural networks follows $\epsilon_r \sim \exp(-\sqrt{n_{iters}})$, which is consistent with that for regression problems (figure \ref{fig:RegmNN64}$c$). ($d$ \& $e$) Comparison of the higher-stage trained network with the error of lower-stage training is shown in the left column. The equation residue $r_n(x)$ for different stages of training is in the middle. Frequency domain of the equation residue $r_n(x)$ at each stage is shown in the right column. ($f$) Prediction error $e_3(x)$ and the equation residue $r_3(x)$ after the third-stage of the training. The zoom-in figure (on the right) shows fluctuations in the prediction error $e_3(x)$, which is caused by the round-off error of the machine-precision of double-floating point. }
\end{figure}

\subsection{Algorithm of multi-stage training for PINNs} \label{sec:mNN_PINN}
Leveraging the multi-stage training algorithm for regression problems and incorporating the results discussed in the previous sections, we have extend the multistage training scheme to physics informed neural networks (PINNs). The details of the algorithm are provided in Algorithm \ref{alg:pinn}.

\begin{algorithm}
	\caption{Multi-stage training scheme for PINNs}\label{alg:pinn}
	\begin{algorithmic}[1]
		\item[] {\bf Prerequisite}: normalize the equation with the magnitude of the largest term around $O(1)$. \vspace{0.5em}
		\item Building the first neural network $u_0(x)$ using regular weight initialization. \vspace{0.5em}
		\item Sampling collocation points or data points e(i.e. boundary conditions) and training the neural network. \vspace{0.5em}
		\item Calculate the output of the trained neural network $u_0(x)$ and corresponding equation residue $r_1(x,u_0)$. \vspace{0.5em}
        \item Estimate the scale factor ${\kappa}_1$, and magnitude prefactor $\epsilon_1$ of the prediction error $e_1$ for the first trained network $u_0$, based on the dominant frequency $f_d^{(r)}$ and magnitude $\epsilon_{r_1}$ of the equation residue $r_1(x, u_0)$, using the relations \eqref{eq:fd2} and \eqref{eq:mag2} or Algorithm \ref{alg:epsil}. \vspace{0.5em}
        \item Generate the ansatz of the solution for the second stage of training as
        \begin{equation}\label{eq:ansz2}
            u_1^{(c)} = u_0(x) + \epsilon_1 u_1(x, \kappa_1) \, ,
        \end{equation}
        where $u_1(x, \kappa_1)$ represents the second neural network utilizing the $sin(x)$ activation function in the first hidden layers and multiplying the scale factor $\kappa_1$ from Step 4 to the weight between the input layer and first hidden layer. we note that $u_1$ has normalized output value.   \vspace{0.5em}
        \item Substituting the ansatz \eqref{eq:ansz2} into the original equation. We note that, in the second stage, only the weight and biases in the network $u_1$ are trainable. The first trained network $u_0$ in the second stage is considered as a known function with fixed parameters. 
        \vspace{0.5em}
        \item  Determine the optimal value of the equation weight $\gamma$ using the relation \eqref{eq:lamb_c} or Algorithm \ref{alg:lambda}. Determine the optimal value of the weight $\gamma_g$ for gradient constraint if the gradient-enhanced PINN is used. \vspace{0.5em}
        \item Conduct the second stage of training and calculate the corresponding equation residue $r_2(x, u_1^{(c)})$.  \vspace{0.5em}
		\item {\bf} Repeat Step 3 - 8 for higher stages of training until the equation residue $r_{n}(x, u_{n-1}^{(c)})$ is small enough. For each higher stage of training, the ansatz of the solution can be expressed in a recursive relation based on the previous trained network as
        \begin{equation}\label{eq:ansz3}
            u_k^{(c)} = u_{k-1}^{(c)}(x) + \epsilon_k u_k(x, \kappa_k) \, ,
        \end{equation}
        where $u_k$ is the new network added at the stage $k$ and $u_{k-1}^{(c)}(x)$ is the combined network from the previous stage of training. $\epsilon_k$ and $\kappa_k$ are the magnitude prefactor and scale factor, respectively, estimated from the equation residue for the previous stage of training in Step 4.
         \vspace{0.5em}

		\item Combining the neural networks from all $n$-stages of training to generate the final solution 
        \begin{equation}\label{eq:ansz4}
            u(x) = u_{n-1}^{(c)} = u_0 + \sum_{k=1}^{n-1}\epsilon_k u_k(x, \kappa_k) \, .
        \end{equation}
        The magnitude of error of the final solution $u(x)$ can be estimated from the equation residue $r_n(x, u_{n-1}^{(c)})$ for the last stage of training by the relation \eqref{eq:mag2} or Algorithm \ref{alg:epsil}. 
	\end{algorithmic}
\end{algorithm}

Here, we note that the primary distinction between the multistage training scheme for PINNs and that for regression problems lies in the fact that we {\it lack} training data for the solution itself for PINNs. Contrasting with the multistage framework for regression problems, where the second network is trained directly using the error $e_1 = u_g - u_0$ between the first trained network $u_0$ with the data $u_g$, we don't necessarily have access to the error of the first trained network in the context of PINNs. Thus, the method of training the second network $u_1$ for PINNs involves creating a combined network $u_{k}^{(c)}$ \eqref{eq:ansz2} that involves the previously trained network $u_{k-1}^{(c)}$ and a new network $u_k(x, \kappa_k)$, with an appropriately-estimated magnitude prefactor $\epsilon_k$ and scale factor $\kappa_k$. A key advantage of this approach is that it circumvents the need to derive a new equation, as shown in \eqref{eq:odegen3}, for each higher-stage network. By fixing the trained weights and biases in the previous networks, the training process for solving the original equation becomes mathematically equivalent to solving the higher-stage governing equation \eqref{eq:odegen3} with the high-frequency source function from the equation residue for the lower-stage training.

Using Algorithm \ref{alg:pinn}, figure \ref{fig:mPINN1d} shows the three-staged PINN training for solving the ordinary differential equation \eqref{eq:eqn1}. For the first two stages, we employ a combination of Adam and L-BFGS for training, which maximizes the convergence rate. However, given the high-frequency residue from the second stage of training, it indicates a high-frequency solution for the third stage of training. Thus, we only use Adam with stochastic gradient descent (SGD) to optimize the performance of the third-stage training, in accordance with the suggestions made in Section \ref{sec:opt}. By combining all the optimal settings as discussed in the previous sections (\S\ref{sec:PINN_mag}-\S\ref{sec:PINN_add}), the prediction error at each stage can be reduced by 3-5 orders of magnitude within $10^5$ iterations.

\begin{figure}[t]
	\centering
	\includegraphics[width=\textwidth]{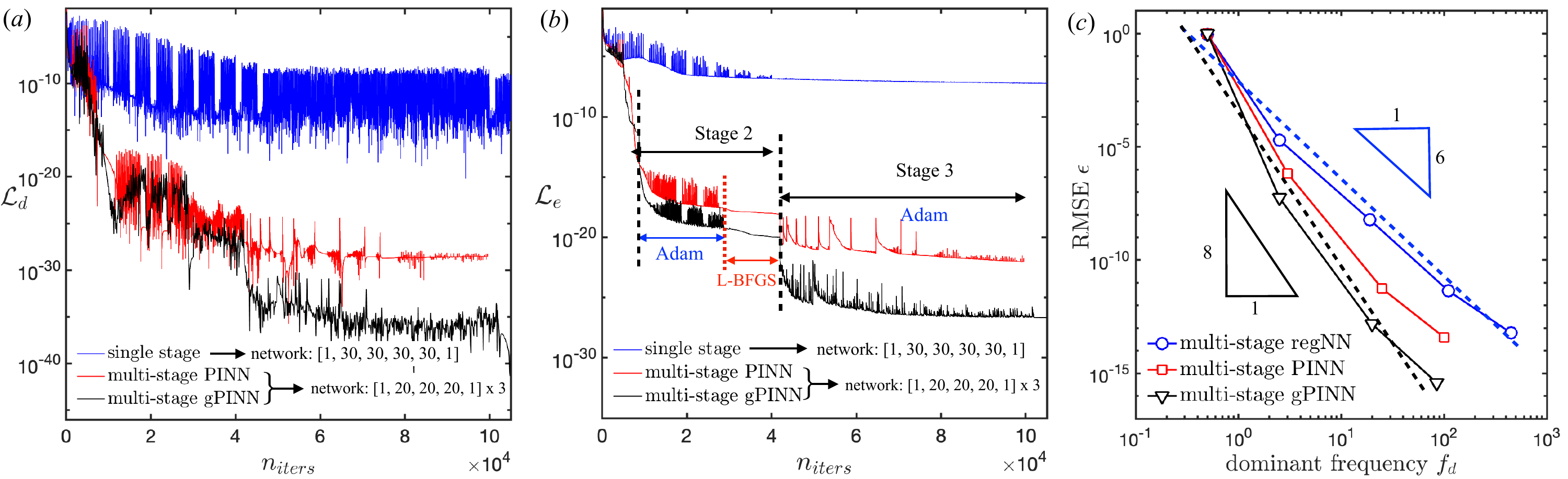}
	\caption{\label{fig:1dcomp} {\bf Comparison of single-stage with multi-stage PINN training}. ($a$ \& $b$) Comparison of the data loss ($a$) and equation loss ($b$) evolution over iterations between the single-stage training, multistage training of PINN, and multistage training of gPINN. ($c$) Relation of the dominant frequency $f_d$ with the root mean square value $\epsilon$ of the error $e_n(x)$ after different stages of training for multi-stage training for regression problems (blue), PINNs (red) and gPINNs (black).}
\end{figure}

Compared with single-stage training, figure \ref{fig:1dcomp}($a$\&$b$) shows that multi-stage training can reduce both the data loss and equation loss by more than 20 orders of magnitude within the same number of iterations. In this instance, the number of weights in the single-stage network has been selected to be approximately equivalent to the total number across all three-stage networks. These results suggest that employing appropriate network settings and an effective training scheme plays a more essential role in successful training than simply increasing the size of neural network and the number of collocation points.  Additionally, when combined with gPINN, the multistage training demonstrates an accelerated convergence rate. Figure \ref{fig:mPINN1d}($f$) shows that, after the three stages of gradient-enhanced PINN training, the prediction error of the final trained network with the exact solution reaches the machine precision of double floating points The observed oscillation in $e_3$ is primary attributable to round-off error. 

\begin{figure}
	\centering
	\includegraphics[width=\textwidth]{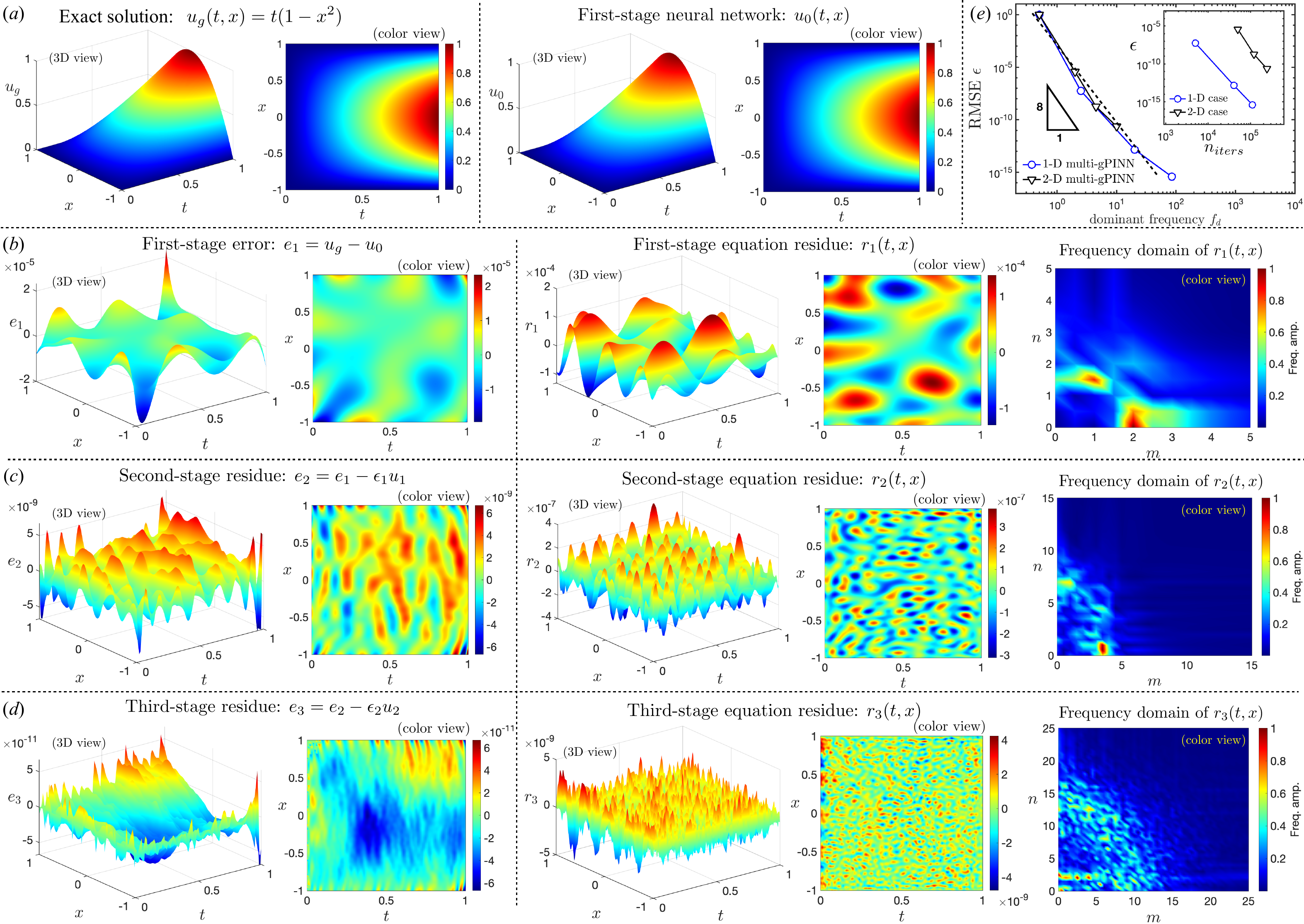}
	\caption{\label{fig:2dcomp} {\bf Multi-stage gPINNs for 2D equations}. ($a$) Comparison of the first-stage trained neural network with the exact solution $u_g(x,t)$ to the equation \eqref{eq:heateqn}. ($b$-$d$) The error $e_n(x,t)$ of higher-stage trained network $u_n(x,t)$ with the exact solution $u_g(x,t)$ is shown in the left panel. The equation residue $r_n(x,t)$ for different stages of training and their frequency domains are shown in the right panel. ($e$) Relation of the dominant frequency $f_d$ with the root mean square value $\epsilon$ of the error $e_n(x,t)$ after different stages of training follows the same power law as 1D problems, of which the exponent $\alpha = 1/8$. The inset shows that the number of iterations required for 2D problems to reach the same accuracy $\epsilon$ is more than that for 1D problems. } \vspace{-0.9em}
\end{figure}

The right panel of figure \ref{fig:mPINN1d}($d$-$f$) displays the spectrum and dominant frequency of the equation residue after each stage of training. Figure \ref{fig:1dcomp}($c$) further shows the relation of the dominant frequency $f_d$ with the root mean square value $\epsilon$ of the prediction error $e_n(x)$ over the stages, which follows a power law $f_d \sim \epsilon^{-\alpha}$ for both regression problems and PINNs.  We recall that the power law exponent $\alpha$ for regression problems is around 1/6.  Compared with that, the power law exponent $\alpha$ for PINNs becomes noticeably smaller, around 1/7 for multi-stage training with regular PINNs and reduced further to 1/8 when using gradient-enhanced PINNs.  As discussed in Section \ref{sec:mNN_reg}, this indicates that trained neural networks in PINNs achieve higher accuracy in capturing higher-order derivatives compared to regression problems. This is reasonable as PINNs involve differential equations that contain the derivatives of the solution. By minimizing the equation loss, PINNs constrain both the neural network and its derivatives to approach the exact solution, enhancing the capture of the high derivative information of the solution. The same reasoning applies to the gradient-enhanced PINNs, which result in an even lower exponent $\alpha$ since the gradient of equation residue involved further higher derivatives of the solution.

\subsection{Application to 2D partial differential equations}
Multi-stage training scheme for PINNs can also be applied to solve partial different equations (PDE). Figure \ref{fig:2dcomp}($a$) shows the three-stage training to solve the diffusion equation
\begin{equation}\label{eq:heateqn}
    \frac{\partial u}{\partial t} = \frac{\partial^2 u}{\partial x^2} + (1-x^2 + t) \qquad \text{with} \ \ u(x, 0) = u(\pm 1, x) = 0 \, ,
\end{equation}
which has the exact solution,
\begin{equation}
    u_g(t,x) = t(1-x^2) \, .
\end{equation}

Consistent with regression problems, the convergence rate of multistage PINN method for solving 2D problem is slightly slower than that for 1D problem (inset of figure \ref{fig:2dcomp}$e$). After three stage of training using RAR method and gradient-enhanced PINN, the prediction error $e_3(t, x)$ of the combined trained networks with the exact solution $u_g(t, x)$ is around $O(10^{-11})$. The accuracy of multistage training is still seven order-of-magnitude higher than that of the single-stage training. Figure \ref{fig:2dcomp}($e$) shows that, when employing the multi-stage training scheme with gPINNs, the relation of the dominant frequency $f_d$ with the root mean square value $\epsilon$ of the prediction error $e(x,y)$ for solving both 1D and 2D equations, follows the same power law \eqref{eq:fde} with an exponent of $\alpha \approx 1/8$. This observation is consistent with the results observed in regression problems (figure \ref{fig:Reg2DmNN}$e$).

We note that achieving a low prediction error of $O(10^{-11})$ for solving 2D partial differential equations via classical numerical method, such as finite difference, would require an extensive number of grid points. For instance, considering the central difference method along the $x$-direction, to reach $10^{-11}$, we would need a grid size in the x-direction of $h_{(x)} \sim O(\sqrt{10^{-11}}) \sim O(10^{-5})$, namely $10^{6}$ grid points for each time step. Even with a 4th-order Runge-Kutta method along the $t$ direction, the step size in $t$-direction would need to be $h_{(x)} \sim O(\sqrt{(10^{-11}})^{1/4} \sim O(10^{-3})$, requiring $10^3$ time steps. Consequently, the total number of grid points needed to achieve this accuracy across the entire domain would be on the order of $O(10^{9})$. 

In contrast, our approach utilizes fully connected neural network with 4-hidden layers consisting of 30 units for each stage of the training. Thus, the total number of weights and biases used to express the solution is only around $3\times 4 \times 30^2 \approx 10^4$, which is five order of magnitude less than the number of grid points used in a discretized solution. This demonstrates the efficiency and effectiveness of the multi-stage PINN in achieving accurate solution with significantly fewer parameters compared to classical numerical methods. 

\section{Generalization of multistage PINN to a combined forward-and-inverse problem}\label{sec:combPINN}
Here we investigate a specific class of problems in mathematics that requires solving equations (forward problem) and simultaneously inferring unknown parameters in the equation (inverse problem) with a high demand for accuracy, for example, finding self-similar blow-up solutions for nonlinear fluid equations \cite{wang2022self}. The physical significance of the problem was explained in Eggers (2015) \cite{eggers2015singularities}, and a prior study \cite{wang2022self} has developed the implementation of PINNs to solve it. In these problems, the multistage PINN method can play a critical role in achieving accurate results.

Here we focus on the 1D inviscid Burgers' equation for which we know the exact solutions. In \ref{sec:AppD}, we provide a summary of the background knowledge and PINN implementation. The task for the PINN involves discovering the {\it smooth} solution to the nonlinear self-similar Burgers' equation,
\begin{equation}\label{eq:burg_self2}
 -\lambda U+[(1+\lambda)y+ U]\partial_y U=0, \qquad \text{with} \quad U(-2) = 1\, ,
\end{equation}
where the solution $U$ should be an odd function with respect to the independent variable $y$, and $\lambda$ is the unknown parameter to be predicted by PINNs. Smooth solutions to \eqref{eq:burg_self2} exist when $\lambda = 1/(2+2i)$ for $i = 1,2,3... $. For other $\lambda$ values, the solution is non-smooth, exhibiting discontinuity at certain order of its derivatives at the origin $y=0$. Finding the smooth solution with the correct value of $\lambda$ numerically is challenging.

To address the issue, prior study \cite{wang2022self} leveraged PINNs and introduced an additional {\it smoothness} constraint into the loss function, which penalizes the higher-order derivative of the equation residue around the non-smooth position. We note that the minimal order of derivative needed for the smoothness constraint depends on specific problems. In general, it should be larger than the order of smoothness for the non-smooth solution (see \ref{sec:AppD}). Any higher derivative with order larger than the minimal value can be involved in the smoothness constraint as long as it remains computationally feasible. Here we focus on the first smooth solution of the self-similar Burgers' equation \eqref{eq:burg_self2}. The loss function can be expressed as
\begin{gather}\label{eq:loss_bgf}
    \mathcal{L} = (1-\gamma)\mathcal{L}_d + \gamma(\mathcal{L}_e + \gamma_g\mathcal{L}_g) + \gamma_s \mathcal{L}_s \qquad \text{with}  \\
	\mathcal{L}_d = (U(y=-2) - 1)^2    \qquad \text{and} \qquad  \mathcal{L}_e = \frac{1}{N_c}\sum_{i=1}^{N_c} \left|r \left(y_i,  U(y_i ) \right) \right|^2 \\ \label{eq:smooth}
    \mathcal{L}_g = \frac{1}{N_c}\sum_{i=1}^{N_c} \left| \frac{\partial r}{\partial y} \left(y_i,  U(y_i ) \right) \right|^2   \qquad \text{and} \qquad  \mathcal{L}_s = \frac{1}{N_s}\sum_{j=1}^{N_s} \left| \frac{\partial^3 r}{\partial y^3} \left(y_j,  U(y_j ) \right) \right|^2 \\
	\text{with} \qquad  r(y, U) = -\lambda U+[(1+\lambda)y+ U]\partial_y U
\end{gather} 
where $\mathcal{L}_d$ and $\mathcal{L}_e$ are the data loss and equation loss, respectively. $\mathcal{L}_g$ is the gPINNs implementation, which involves the first-order gradient of the equation residue $r(y, U)$. $\mathcal{L}_s$ is the smoothness constraint that incorporates the third derivative of the equation residue. While the equation loss $\mathcal{L}_e$ and gradient loss $\mathcal{L}_g$ are examined at $N_c$ random collocation points $y_i$ across the entire domain, the smoothness constraint $\mathcal{L}_s$ is calculated at $y_j$ close to the origin (e.g. $|y_j|<0.1$) with number $N_s \ll N_c$. Although the smoothness constraint depends on the equation residue, it can be viewed as an additional boundary condition for the solution to determine the value of $\lambda$.  

\begin{figure}[!ht]
	\centering
	\includegraphics[width=\textwidth]{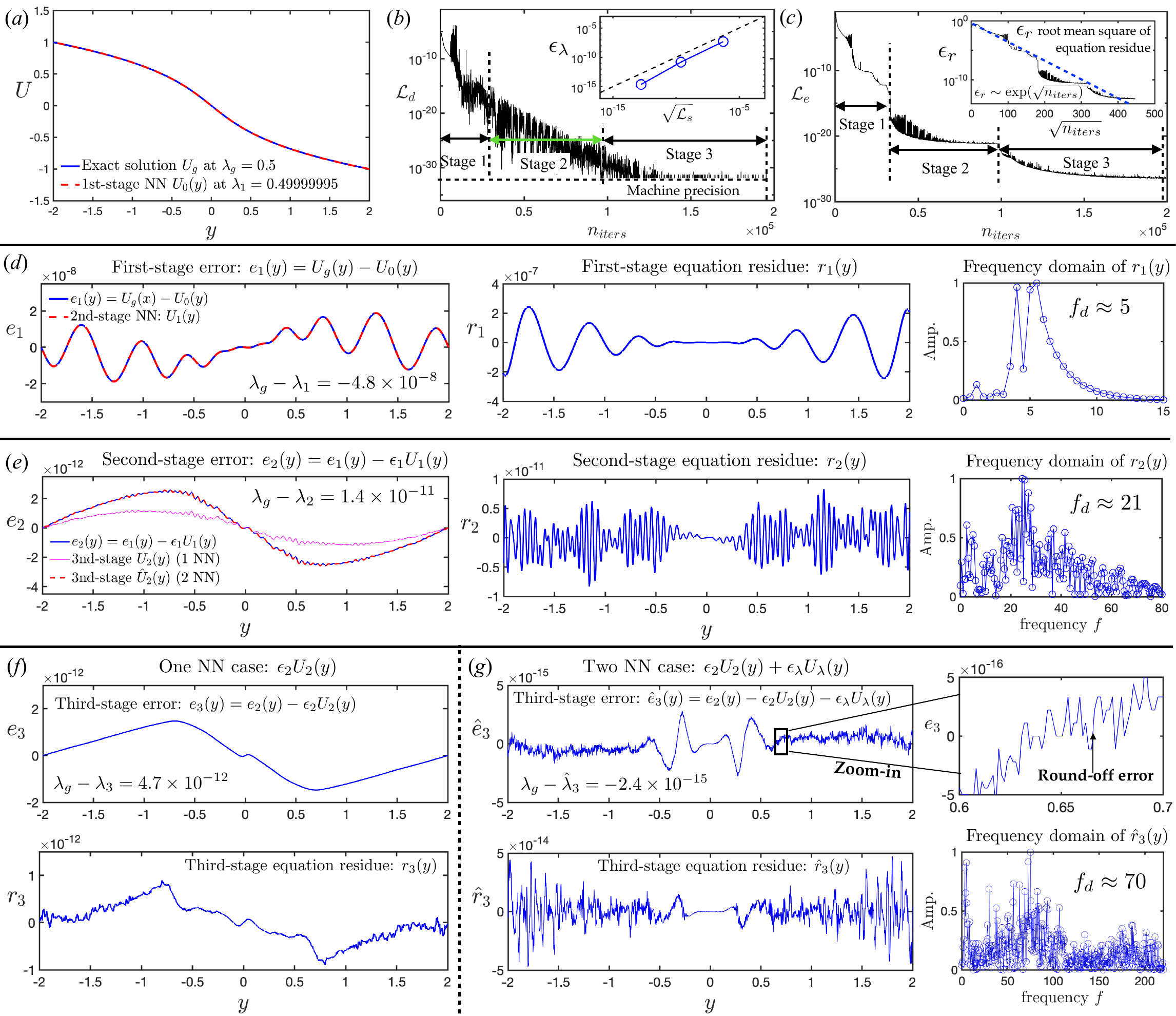}
	\caption{\label{fig:burg_s} {\bf Multi-stage gPINNs for a combined forward and inverse problem}. ($a$) Comparison of the first-stage trained network $U_0(y)$ at the inferred $\lambda_1$ (red dashed curve) with the exact profile of the first smooth solution $U_g(y)$ with $\lambda_g = 0.5$ to the self-similar Burgers' equation \eqref{eq:burg_self2}.  ($b$) Data loss and ($c$) equation loss over iterations of three-stage training. The inset of ($b$) shows the relation of the error $\epsilon_\lambda$ of inferred $\lambda$ with the loss of the smoothness constraint $\mathcal{L}_s$ after different stages of training. The dash line indicates the relation $\epsilon = \sqrt{\mathcal{L_s}}$.  The inset of ($c$) shows that the evolution of the root mean square value $\epsilon_r$ of the equation residue $r(y,u)$ over iterations of the multi-stage neural networks follows $\epsilon_r \sim \exp(-\sqrt{n_{iters}})$, consistent with that of regular forward problem (figure \ref{fig:mPINN1d}$c$). ($d$ \& $e$)  The prediction error $e_n(y)$ (left), equation residue $r_n(y)$ (right) and its frequency domain (right) for the first ($d$) and second ($e$) stages of training. Comparison of higher-stage trained networks with the lower-stage prediction error is shown in the left panel.  ($f$) Prediction error $e_3(y)$ and equation residue $r_3(y)$ for the third-stage training using only one additional neural network $U_3(y)$. It successfully reduce the high-frequency error from the second stage but fails to reduce its low-frequency error. ($g$) Prediction error $\hat{e}_3(y)$ and equation residue $\hat{r}_3(y)$ for the third-stage training using two neural networks $U_3(y)$ and $U_\lambda(y)$, which successfully reduce both the high-frequency error associated with lower-stage equation residue $r_2(y)$ and the low-frequency error associated with the error $\epsilon_\lambda$ of inferred $\lambda_2$. The zoom-in figure shows the prediction after three stages of training approaches the machine precision of double floating point.}
\end{figure}

Following Algorithm \ref{alg:pinn}, figure \ref{fig:burg_s}($a$-$d$) shows the first two stages of training for solving the self-similar Burgers' equation \eqref{eq:burg_self2}. We observe that the second-stage training successfully improves both the prediction error $e_2(y)$ of the trained network and the inferred lambda $\lambda_2$ by \textit{four orders of magnitude}. However, in addition to the high-frequency error as previously seen for the higher-stages, we observed that the prediction error $e_2(y)$ from the second-stage training contains a low-frequency profile, which dominates over the high-frequency error. This disparity hinders the further reduction of the error by adding more stages of training based on Algorithm \ref{alg:pinn}.

To understand the issue, we first study the occurrence of the high-frequency error in $e_2(y)$. The middle panel in figure \ref{fig:burg_s}($e$) reveals that the equation residue $r_2(y)$ after the second stage of training exhibits a similar dominant frequency $f_d$ to the high-frequency error in the prediction error $e_2(y)$. Using \eqref{eq:err_mag}, we estimate the magnitude $\epsilon_2$ of the prediction error $e_2(y)$ based on the magnitude $\epsilon_{r_2}$ of the equation residue $r_2(y)$ as $\epsilon_2 = \epsilon_{r_2}/(2\pi f_d) \sim O(10^{-13})$. This is consistent with the magnitude of the high-frequency error in $e_2(y)$.  Then, following Algorithm \ref{alg:pinn}, we create a new network $U_2(y)$ multiplied by the magnitude of $O(10^{-13})$ for the third-stage training. Figure \ref{fig:burg_s}($f$) shows that the high-frequency error in $e_2(y)$ after the third-stage training does vanish in $e_3(y)$. However, the magnitude of the prediction error $e_3(y)$ and the inferred $\lambda_3$ after the third-stage of training (figure \ref{fig:burg_s}$f$) remains nearly the same as those of the second-stage training (figure \ref{fig:burg_s}$e$). 

The issue appears to be related to the existence of the low-frequency profile in $e_2(y)$.  We recall that the prediction error of the training is estimated by comparing the trained networks at the inferred $\lambda_2$ with the exact smooth solution at $\lambda_g = 0.5$. Therefore, the error of the trained network is influenced not only by the equation residue, but also by the error $\epsilon_\lambda$ of the inferred $\lambda$. To assess the impact of the inference error $\epsilon_\lambda$ on the prediction error $e_2(y)$, we perform a similar analysis as discussed in Section \ref{sec:simple}, introducing the ansatz of the exact solution $U_g$ and exact value of $\lambda_g$ as
\begin{eqnarray}\label{eq:ansatz_b}
	U_g(y) = U_0(y) + \epsilon U_e(y) \qquad \text{and} \quad \lambda_g = \lambda_0 + \epsilon_\lambda
\end{eqnarray}
where $U_0$ represents the lower-stage trained network and $\lambda_0$ is the inferred $\lambda$ from the lower-stage training.  $\epsilon U_e(y) $ represents the prediction error of the trained network and $\epsilon_\lambda$ is the error of inferred $\lambda$. Both $\epsilon$ and $\epsilon_\lambda$ are much smaller than 1. Substituting \eqref{eq:ansatz_b} into \eqref{eq:burg_self2} and removing higher-order small terms $O(\epsilon^2)$, we have
\begin{equation}\label{eq:burg_stage3}
	\epsilon \left\{ (\partial_y U_0-\lambda_0) U_e+[(1+\lambda_0)y+ U_0]\partial_y U_e \right\} = \underbrace{(\lambda U_0-[(1+\lambda_0)y+ U_0]\partial_y U_0)}_{\text{equation residue: }-r_0} + \underbrace{\epsilon_\lambda (U_0 - y \partial_y U_0)}_{\text{term from }\epsilon_\lambda: \: r_\lambda}
\end{equation}
which can be viewed as the governing equation for the higher-stage network. In addition to the equation residue $r_0(y)$ from the lower-stage training,  the higher-stage equation \eqref{eq:burg_stage3} involves a new source function $r_\lambda(y)$ that is associated with the error $\epsilon_\lambda$ of the inferred $\lambda$. While the equation residue $r_0(y)$ exhibits high-frequency behavior, the source function $r_\lambda(y)$ is influenced by the profile of the trained network $U_0(y)$, exhibiting the low-frequency profile in the prediction error $e_2(y)$ (figure \ref{fig:burg_s}$e$). 

Considering the low frequency nature of the source function $r_\lambda(y)$, the magnitude of prediction error $\epsilon$ in \eqref{eq:burg_stage3} associated with $r_\lambda(y)$ is expected to be similar to the error $\epsilon_\lambda$ of the inferred $\lambda$, which is approximately $O(10^{-12})$, consistent with our results (figure \ref{fig:burg_s}$e$). In contrast, the prediction error associated with the high-frequency equation residue $r_0$, as discussed earlier, is only around $O(10^{-13})$. This explains why the low-frequency profile dominates the prediction error $e_2(y)$. 

Here, we note that the error $\epsilon_\lambda$ of inferred $\lambda$ is calculated using the known exact value $\lambda_g=0.5$. However, in many other problems, the exact value of $\lambda_g$ is unknown. Thus, an alternative way to quantify the inference error $\epsilon_\lambda$ is from the loss $\mathcal{L}_s$ of the smoothness constraint.  The inset of figure \ref{fig:burg_s}($b$) shows that the inference error $\epsilon_\lambda$ after different stages of training is proportional to $\sqrt{L_s}$, i.e. $\epsilon_\lambda = \sqrt{L_s}$ (dashed line). This suggests that we can use $\sqrt{L_s}$ to estimate the inference error $\epsilon_\lambda$, as well as the magnitude prefactor for the higher-stage network $U_\lambda$ associated with $r_\lambda(y)$.

Since the prediction error is dominated by the low-frequency source function $r_\lambda(y)$, one might intuitively consider creating a single low-frequency network multiplied by the error $\epsilon_\lambda$ of the inferred $\lambda$ for the third-stage training. However, this approach is not effective because the smoothness constraint \eqref{eq:smooth} depends on the higher-order derivative of the equation residue. By using only a low-frequency network, it would be challenging to reduce the high-frequency equation residue. Therefore, our proposed solution is to create two networks for both source functions in \eqref{eq:burg_stage3} at the third-stage training, namely 
\begin{equation}
    \epsilon_2 U_2(y) +  \epsilon_\lambda U_\lambda(y),
\end{equation}
where the magnitude prefactor $\epsilon_2$ and modified scale factor $\hat{\kappa}$ (for frequency) for the high-frequency network $U_2(y)$ associated with the equation residue $r_2$ can be determined by the relations \eqref{eq:mag} and \eqref{eq:fd} or Algorithm \ref{alg:epsil}. The low-frequency network $U_\lambda$ associated with the error $\epsilon_\lambda$ of the inferred $\lambda_2$ can be directly multiplied by the inference error $\epsilon_\lambda$. Figure \ref{fig:burg_s}($g$) shows that, using combined two networks for the third-stage training, the prediction error $\hat{e}_3(y)$ is successfully reduced by another three orders of magnitude, eventually approaching the machine precision of double-floating points. 

\section{Discussion}\label{sec:dissc}
We note that the principle of multi-stage neural networks is similar to that of Fourier series, which combines a series of sine or cosine functions, ranging from low to high frequencies, to approximate functions.  Provided the series converge, the error between the Fourier series expansion of a given order and the target function possesses lower magnitudes but higher frequency than any terms in the series. To further minimize the error,  higher-order sine or cosine functions need to be incorporated into the series, leading to additional higher-frequency error. 

The introduction of new neural networks in multi-stage neural networks (MSNNs) is analogous to the inclusion of higher-order trigonometric functions in the Fourier series expansions. However, in contrast to sines and cosines, deep neural networks with appropriate settings offer stronger function representation capacity. Our finding indicates that the magnitude $\epsilon$ of error after each stage of training follows a inverse power law relation with the dominant frequency $f_d$ of the error, i.e. $\epsilon \sim f_d^{-\alpha}$, with the exponent $\alpha \approx 1/6$ for regression problems, $\alpha \approx 1/7$ for regular PINNs, and $\alpha \approx 1/8$ for gradient-enhanced PINNs (gPINNs).

In comparison, the power law exponent $\alpha$ for Fourier series is roughly around $\alpha \approx 0.5$ \cite{kreyszig2007advanced}, much larger than that of MSNNs. This indicates that, to achieve the same error magnitude, the error frequency generated by MSNNs could be several orders of magnitude lower than that by Fourier series. This observation confirms that MSNNs serve as a superior tool capable of accurately approximating target functions, as well as their high derivative information.

The multi-stage neural networks (MSNNs) developed in this work remains in their {\it early} stage, and mainly serve as a proof of concept to demonstrate that neural networks can practically achieve high accuracy. It is crucial to recognize that MSNNs should not be regarded as a substitute for classical numerical methods, but rather as a complementary approach. In fact, there remains several challenges that need to be addressed in the MSNN method. One of the primary challenges pertains to high-dimension problems. As shown in figure \ref{fig:Reg2DmNN} and \ref{fig:2dcomp}, the convergence rate of MSNNs for both 2D regression problems and PINNs are consistently slower than that for 1D problems. It is expected that this challenges will become more pronounced in higher-dimensional problems. 

The second major challenge pertains to approximating functions or predict solutions with steep gradients. Near the regions where the target function exhibits steep gradients, neural networks often encounter local peaks in the error or the equation residue during training. The presence of these peaks hinder the reduction of error in successive stages, necessitating their removal before proceeding to the next stage of training. We note that functions with steep gradients are commonly encountered in differential equations, such as stiff equations, nonlinear equations, or singular perturbation equations (see \ref{sec:AppC}). Solving these types of equation via PINNs is beyond the scope of this paper.

There are additional questions to be addressed that could further improve the MSNN method. One of the critical questions concerns the optimal timing for transitioning to the next stage of training. In each stage, the convergence rate of training loss gradually decreases over the iterations. The decision whether to switch to the next stage quickly for higher convergence rates, or to stay in the current stage until the loss plateaus in order to maximize the error reduction at each stage requires careful consideration and further investigation. Moreover, with multiple stages of training, oversized networks are no longer required to achieve high accuracy within a single stage of training. The optimal strategy for selecting the neural network size at different stages that can minimize the numbers of training parameters (weights and biases) and thus computational expense for the entire MSNN training becomes another future direction that is worth investigating.

\section{Conclusion}\label{sec:concl}
We introduced the multi-stage neural networks (MSNNs) for both regression problems and physics-informed neural networks. Inspired by perturbation theory, we sequentially introduced new stages of training with new neural networks to capture the residue from the previous stage of training. This enable MSNNs to reach unprecedentedly high accuracy over stages. We showed that three stages of MSNN training can reach machine precision, making neural networks truly universal function approximators in practice. This new method can be widely applied to many scientific domains, such as mathematical and nonlinear physical science where the precision matters.

The success of MSNNs lies in two aspects. The first is the idea of staged training itself. Deep neural networks often suffer from spectral biases, making it challenging to capture the full spectrum of the target function in a single stage of training, even when employing large-sized networks with an increased number of data or collocation points. As a result, the training loss tends to plateau after a certain number of iterations. However, by employing multi-stage training, the previously plateaued error can be substantially reduced in each successive stage, which enables MSNNs to progressively capture finer details of the target function. 

The second aspect for the success of MSNNs is the specific design of each new-stage network based on the training error from the previous stage. The neural-network predictions in successive stages exhibits significantly small magnitude and high frequency compared to the previous stages. We showed that, by employing optimal magnitude prefactor $\epsilon$ and scale factor $\kappa$ with \texttt{sin} activation function, accurate predictions of functions with small magnitudes and high frequencies can be achieved. This enables the effective capture of intricate features in each successive stage. 

To maximize the performance of each stage of training, we also studied the optimal value of $\epsilon_n$ and $\kappa_n$ for each stage. For regression problems, the $\epsilon_n$ is equal to the magnitude (root mean square value) of the error $e_n({\bf x})$ between the trained networks in the previous stage and the ground truth $u_g({\bf x})$, and $\kappa_n$ is proportional to the dominant frequency $f_d$ of the error $e_n({\bf x})$. 

However, for physics-informed neural networks (PINNs), the prediction error $e_n({\bf x})$ is not directly available and needs to be inferred from the equation residue $r_n({\bf x})$ from the previous stage of training. Based on the fact that the governing equations for higher-stage training are essentially linear, we provided the theoretical relations between the magnitude and frequency of the prediction error $e_n({\bf x})$ and the equation residue $r_n({\bf x})$ in Section \ref{sec:PINN}. We also presented an algorithm that can effectively estimate the magnitude of the prediction error from the equation residue.

Moreover, we discussed several other optimal settings that can enhance the efficiency of multi-stage PINN training. These include the equation weight $\gamma$, number of collocation points $N_c$, choice of optimizer, and advanced PINN techniques in the literature, such as RAR method and gPINNs.

Leveraging all the optimal settings discussed in this work, we showed that multi-stage neural networks (MSNNs) can significantly reduce the prediction error for both regression problems and PINNs, approaching machine precision. Furthermore,  MSNNs showcases their capability in solving combined-forward-and-inverse problems to machine precision, a task typically challenging for classical numerical methods, but of great importance in mathematical and physical sciences. However, there remains many questions and challenges to be addressed for further enhancing the MSNN method. 

\section*{Acknowledgements}
We thank T. Buckmaster and J. G{\'o}mez-Serrano for helpful discussions regarding the application of multi-stage neural networks to critical mathematical questions. We acknowledge the Office of the Dean for Research at Princeton University for partial funding support via the Dean for Research Fund for New Ideas in the Natural Sciences. C.-Y.L acknowledge the National Science Foundation for funding via Grant No. DMS-2245228. We also gratefully acknowledges financial support from the Schmidt Data X Fund at Princeton University made possible through a major gift from the Schmidt Futures Foundation.

\setcounter{section}{0}
\renewcommand\thesection{Appendix \Alph{section}}
\renewcommand\theequation{\Alph{section}.\arabic{equation}}

\section{Neural network error under different settings}\label{sec:AppA} \vspace{-0em}
\begin{figure}[H]
	\centering
	\includegraphics[width=1\textwidth]{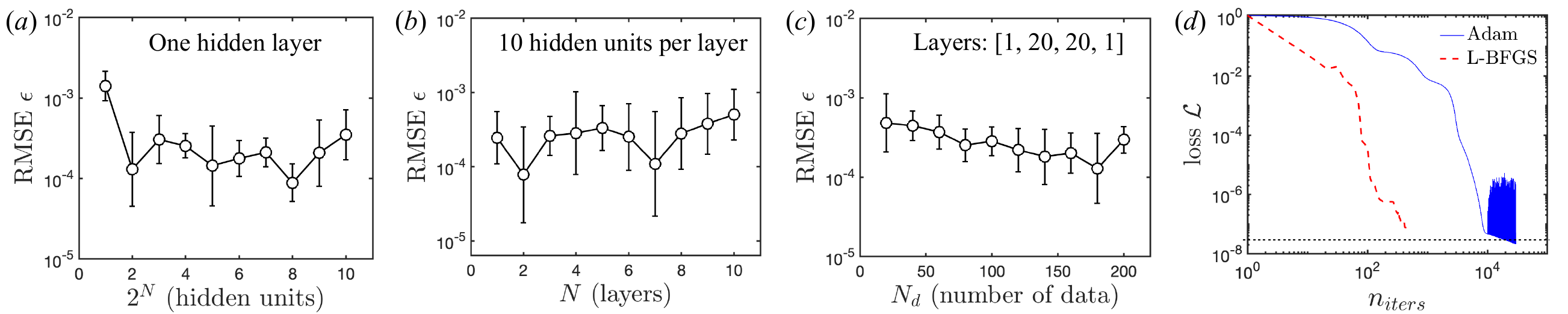}
	\caption{\label{fig:sysexp}  Root mean square (RMS) value $\epsilon$ of the error between the target function and trained network using different number of ($a$) hidden units,  ($b$) layers, and ($c$) training data, and ($d$) different types of optimizers, which show no big difference. Error bars show the standard deviation of eight repetitive experiments with different random initialization.} \vspace{-0em}
\end{figure}
Systematic experiments (figure \ref{fig:sysexp}) show that the root mean square value (RMS) $\epsilon$ of the error $e(x)$ between the trained network $u_0(x)$ and the data from the target function $u_g(x)$ remains unchanged even when the number of either layers (figure \ref{fig:sysexp}$a$) or units (figure \ref{fig:sysexp}$b$) is increased. Although the RMS error $\epsilon$ does slightly decrease with an increase in training data (figure \ref{fig:sysexp}$c$), this reduction is smaller than the standard deviation of eight repetitive experiments conducted with different random initializations, and thus is negligible. These results suggest that the plateau in training loss is not due to insufficient neural network size or lack of training data, but instead arises from inherent limitations of the training process itself. Figure \ref{fig:sysexp}($d$) presents the training loss for two different optimization methods. Compared to Adam \cite{kingma2014adam}, a first-order gradient descent method, L-BFGS \cite{liu1989limited}, a quasi-Newton method, exhibits a higher convergence rate. However, training with L-BFGS quickly falls into a local minimum after reaching the same plateau as Adam. This suggests that the loss plateau is not optimizer-specific.

\vspace{1em}

\section{Effect of data magnitude on neural network training}\label{sec:AppB} \vspace{-0em}
\begin{figure}[H]
	\centering
	\includegraphics[width=1\textwidth]{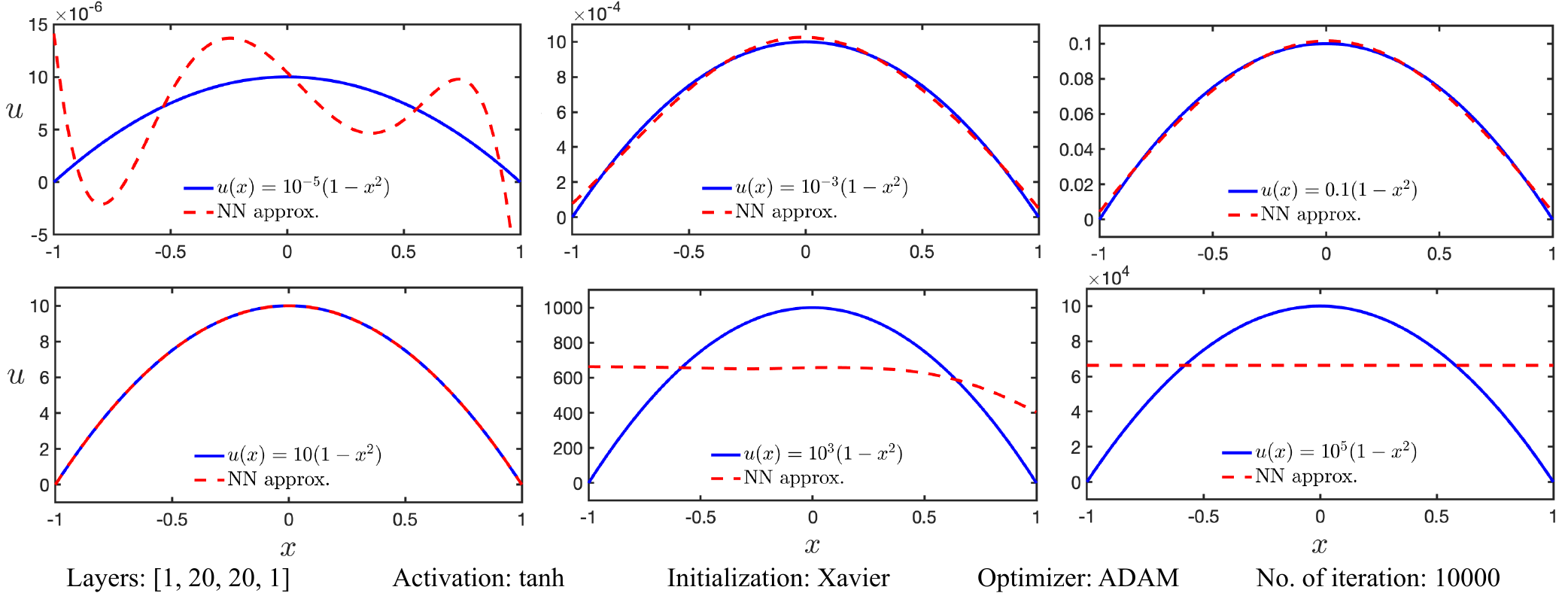}
	\caption{\label{fig:scale1}  Fitting of neural networks to the data with different magnitudes without normalization. It shows that the network is hard to fit data with magnitude either too much larger or smaller than 1.} \vspace{-0em}
\end{figure}
\vspace{1em}

\section{Two extreme types of equations}\label{sec:AppC}
There are two extreme types of equation where the general settings of networks derived in \eqref{eq:mag2} and \eqref{eq:fd2} for the high-stage training do not strictly hold. The {\it first} case is when the equation involves nonlinear term with high-order of derivatives, for example,
\begin{equation}\label{eq:expfail1}
	\left(\frac{d^8 u}{dx^8}\right)^2 - u = F(x).
\end{equation}
Substituting the ansatz \eqref{eq:ansatz} into \eqref{eq:expfail1} gives
\begin{equation}\label{eq:expfail2}
		-\epsilon\left( 2 \frac{d^8 u_0}{dx^8} \frac{d^8 u_1}{dx^8} - u_1 \right) - \epsilon^2 \left(\frac{d^8 u_1}{dx^8}\right)^2 = r(x, u_0)  =  \left(\frac{d^8 u}{dx^8}\right)^2 - u_0 - F(x)
\end{equation}
where $r(x, u_0)$ is the equation residue of $u_0$. When $u_1$ is a high-frequency function with dominant frequency $f_d$ satisfying the criterion,
\begin{equation}\label{eq:ncri}
	(2 \pi f_d)^8 \epsilon > 1  \qquad \Longrightarrow \qquad  f_d > \epsilon^{-1/8},
\end{equation}
the nonlinear term of $u_1$ on the left-hand side of \eqref{eq:expfail2} is no longer negligible and becomes the dominant term in the equation. The magnitude and frequency of $u_1(x)$, thus, need to be reassessed by balancing the nonlinear term with the equation residue $r(x, u_0)$. This results in the dominant frequency $f_d^{(1)}$ of $u_1(x)$ to be $f_d^{(1)} = f_d^{(e)}/2$, rather than $f_d^{(1)} = f_d^{(e)}$ from \eqref{eq:fd}, where $f_d^{(e)}$ represents the dominant frequency of the equation residue $r(x, u_0)$.  

However, we note that, although the dominant frequency $f_d^{(e)}$ of equation residue is larger, it still capture the order of magnitude of the actual frequency of $u_1(x)$. We recall from figure \ref{fig:overfit}($c$) that neural networks with modified scale factor $\hat{\kappa}$ larger than the criterion $\hat{\kappa}>\pi f_d$ can reach the same high-accuracy of fitting to high-frequency functions. This indicates that setting the scale factor $\kappa$ based on a larger dominant frequency $f_d^{(e)}$ for the network of $u_1$ remains a good option to solve \eqref{eq:expfail2}. \\

The {\it second} case is  when the equation involves singular perturbation term,  for example
\begin{equation}\label{eq:expfail3}
	\alpha \frac{d^4 u}{dx^4} + \frac{d^2 u}{dx^2}- u = F(x) \qquad \text{with} \quad \alpha \ll 1.
\end{equation}
where the coefficient before the highest-order derivative term is much smaller than the others. This type of equation is very common in physical sciences, such as boundary-layer problem. Substituting the ansatz \eqref{eq:ansatz} into \eqref{eq:expfail3} gives
\begin{equation}\label{eq:expfail4}
-\epsilon\left(\alpha \frac{d^4 u_1}{dx^4} + \frac{d^2 u_1}{dx^2}- u_1 \right) = r(x, u_0) = \alpha \frac{d^4 u_0}{dx^4} + \frac{d^2 u_0}{dx^2}- u_0-F(x)
\end{equation}
where $r(x, u_0)$ is the equation residue of $u_0$. Based on \eqref{eq:expfail4}, the dominant frequency $f_d^{(1)}$ of $u_1(x)$ remains equal to that of the equation residue $r(x, u_0)$. However, when $\alpha < (2\pi f_d)^{-2}$, the dominant term on the right-hand side of \eqref{eq:expfail4} is not the one with the highest-order derivative of $u_1$, but the term with the second-order derivative.  The magnitude $\epsilon$ of the error is, then, determined by
\begin{gather}\label{eq:mag3}
	\epsilon = \frac{\epsilon_r}{\left[2\pi f_d^{(1)}\right]^2},  \qquad \text{rather than }  \ \ \frac{\epsilon_f}{\left[2\pi f_d^{(1)}\right]^4 \alpha},
\end{gather}
which is based on \eqref{eq:mag}. Here, we note that the actual challenges of solving singular perturbation equation via PINNs is more than the violation of the expression \eqref{eq:mag} for setting the higher-stage neural network. According to asymptotic analysis, the existence of singular perturbation term in the equation indicates that the solution to the equation has a narrow inner region where local gradient is very large. This property makes both the first-stage and higher-stage training of networks difficult. More discussion of the challenge is given in the Discussion (Section \ref{sec:concl}) of the paper. However, the solution to this challenge is beyond the scope of this paper.

\section{1D inviscid Burgers' equation}\label{sec:AppD}

This section summarizes the exact self-similar blow-up solutions to the 1D inviscid Burgers' equation \cite{eggers2015singularities} and the PINN implementation developed in Wang {\it et al.} \cite{wang2022self} to find it numerically. 

Without viscous dissipation, the 1D inviscid Burgers' equation is given as
\begin{equation}\label{eq:burg_s}
    \frac{\partial u}{\partial t} + u\frac{\partial u}{\partial x} = 0.
\end{equation}
which has a shock wave solution where the velocity becomes discontinuous at a finite time, exhibiting a singularity where the solution blow up. However, right before the time when the shock/singularity is formed, the velocity profile remains smooth and continuous, and follow a self-similar structure near the singularity formation. We suppose the singularity occurs at $t = t_0$ and $x = x_0$. The self-similar coordinates can be written as
\begin{eqnarray}\label{eq:loccd}
	s=-\log(t_0-t),  \qquad  y  =\frac{x-x_0}{(t_0-t)^{1+\lambda}},
\end{eqnarray}
where $s$ and $y$ are the local time and spatial coordinates, respectively. When $s$ goes to infinity, time $t$ approaches to the blow-up time $t_0$, but can never go beyond that. In the meantime, the new self similar coordinate $y$ allow us to zoom into and examine the solution profile around the singularity as time approaches $t_0$. The solution $u$ follows the ansatz \cite{wang2022self}
\begin{equation}\label{eq:u_b}
	u(x,t) = (t-t_0)^\beta U(y,s)
\end{equation}
where $U(y,s)$ indicates the self-similar profile near the singularity with $\beta$ to be determined. Substituting the ansatz \eqref{eq:u_b} into the equation \eqref{eq:burg_s} gives $\beta = \lambda$. Thus, the self-similar form of the Burgers' equation becomes
\begin{equation}\label{eq:burg_self}
	(\partial_s-1)\lambda U+[(1+\lambda)y+ U]\partial_y U=0.
\end{equation}
We assume that when approaching the blow-up time $t_0$, namely $s$ goes to infinity, time derivative term in \eqref{eq:burg_self} vanishes, and the self-similar profile $U(y,s)$ reaches steady state. Then, the steady state profile $\tilde{U}(y)$ is governed by
\begin{equation}\label{eq:burg_self3}
 -\lambda \tilde{U}+[(1+\lambda)y+ \tilde{U}]\partial_y \tilde{U}=0.
\end{equation}
 For simplicity, we consistently use $U$ to represent the steady state solution in the rest of the section. The parameter $\lambda$ \eqref{eq:loccd}, the rate at which singularity forms, remains unknown and is the key parameter to be inferred via the multi-stage neural networks. 

\begin{figure}
	\centering
	\includegraphics[width=\textwidth]{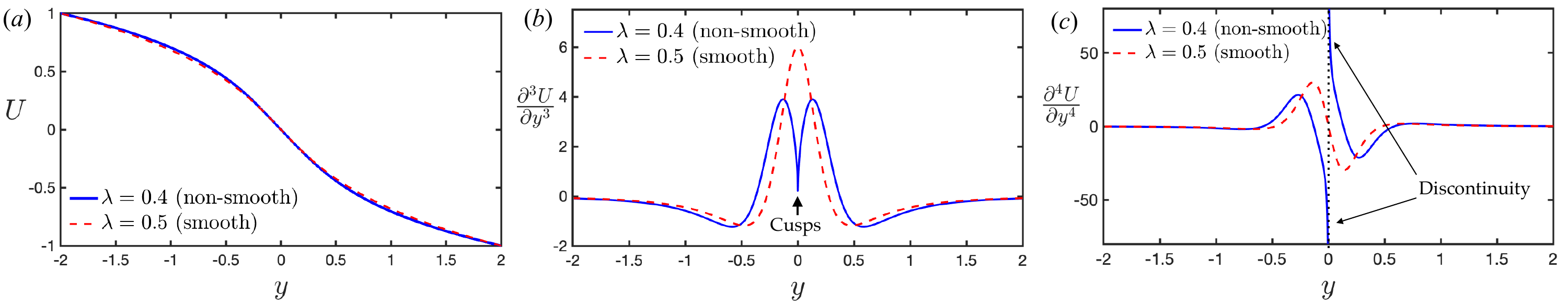}
	\caption{\label{fig:burg_a} {\bf Exact solutions to the Burgers' equation}. ($a$) Exact first smooth and nearby non-smooth solutions to the self-similar Burgers' equation \eqref{eq:burg_self3}. ($b$) the third derivative of the non-smooth solution ($\lambda = 0.4$) forms cusps at the origin which indicates its fourth derivative ($c$) at the origin become discontinuous. In comparison, the smooth solution  ($\lambda = 0.5$) is continuous at all orders of derivative everywhere in the domain. }
\end{figure}

To guarantee the equation \eqref{eq:burg_self3} is well-posed globally in the local coordinates, the self-similar solution $U$ must be an odd function. Theoretically, there exists solutions to \eqref{eq:burg_self3} for each value of $\lambda$. The analytic solutions to the self-similar Burgers' equation \eqref{eq:burg_self3} are
\begin{equation}\label{eq:bganal}
    y= \begin{dcases}
    -U-CU^{1+\frac 1\lambda} \qquad & \text{for} \quad x\geq 0 \\
    -U+C(-U)^{1+\frac 1\lambda} \qquad & \text{for} \quad x< 0
    \end{dcases}
\end{equation}
where $C$ is a constant determined by the boundary condition. Here, we use $U(y=2) =-1$, which gives $C = 1$.  From the analytic expression \eqref{eq:bganal}, we can see that $\lambda$, in fact, determines the smoothness of the solution. Here, the smoothness indicates the solution is continuous at all its derivative. When $\lambda = 1/(2+2i)$ with $i = 0, 1, 2, ...$, the solution is smooth everywhere in the domain. However, when $\lambda \neq 1/(2+2i)$, the expression \eqref{eq:bganal} involves fractional power, causing the solution to be non-smooth at the origin. For example, figure \ref{fig:burg_a}($c$) shows that the fourth derivative of the solution for $\lambda = 0.4$ is discontinuous at the origin. Here, we note that the non-smooth solutions have no physical meaning. Thus, finding the smooth solutions to \eqref{eq:burg_self2} is the goal.

Prior study by Wang {\it et al.} \cite{wang2022self}, leveraged the continuous property of neural networks, showing that PINN can discover the smooth solution with associated $\lambda$ by imposing the high-order derivative constraint at the non-smooth position, known as the {\it smoothness} constraint.  Additionally, we impose odd symmetry of the solution $U$ by constructing the function form $U = y[\mathrm{NN}_u(y) + \mathrm{NN}_u(-y)]$, where $\mathrm{NN}_u$ indicates a fully-connected neural network created for $U$. The {\it data loss} and {\it equation loss} for solving the Burgers' equation \eqref{eq:burg_self3} are given as
\begin{gather}\label{eq:loss_bgf}
	\mathcal{L}_d = (U(y=-2) - 1)^2    \qquad \text{and} \qquad  \mathcal{L}_e = \frac{1}{N_c}\sum_{i=1}^{N_c} \left|r \left(y_i,  U(y_i ) \right) \right|^2 \\
	\text{with} \qquad  r(y, U) = -\lambda U+((1+\lambda)y+ U)\partial_y U
\end{gather} 
where $y_i$ indicates the random collocation points in the training domain $y\in[-2, 2]$ and $N_c$ is their total number. Here we focus on finding the first smooth solution with known $\lambda_g = 1/2$.  Utilizing the fact that the non-smooth solutions in the neighborhood of $\lambda = 0.5$ has unbounded fourth-order derivative, which appears in the third order of derivative of equation residue, the smoothness constraint is given as
\begin{equation}\label{eq:loss_gb1}
	\mathcal{L}_{s}  = \frac{1}{N_s}\sum_{i=1}^{N_s} \left|\frac{d^3 r}{d y^3} \left(y_i, \: U(y_i ) \right) \right|^2\,.
\end{equation}
where $y_i$ indicates the random collocation points close to the origin (e.g. $|y_i|<0.1$) and $N_s$ is their total number. Although the smoothness constraint depends on the equation residue, it can be simply considered as a boundary condition for the solution that help determine the value of $\lambda$. 

\bibliographystyle{unsrt}

\end{document}